\documentclass{article}

\usepackage{microtype}
\usepackage{graphicx}
\usepackage{booktabs} 

\usepackage[title]{appendix}
\usepackage{mathtools,amsmath,amsthm,amssymb}
\DeclareMathOperator{\E}{\mathbb{E}}
\usepackage{pgf}
\usepackage{wrapfig}
\usepackage{pgfplots}
\pgfplotsset{compat=1.16}
\usepackage{caption}
\usepackage{subcaption}
\usepackage{url}
\usepackage{enumitem} 
\usepackage[numbers]{natbib}
\usepackage{footnote}

\newcommand{\eqnref}[1]{\eqref{#1}}

\usepackage{hyperref}

\usepackage[accepted]{mlsys2021}

\mlsystitlerunning{Pipelined Backpropagation at Scale}

\begin{document}

\twocolumn[
\mlsystitle{Pipelined Backpropagation at Scale: \break Training Large Models without Batches}



\mlsyssetsymbol{equal}{*}

\begin{mlsysauthorlist}
\mlsysauthor{Atli Kosson}{equal,wdwac,tsla}
\mlsysauthor{Vitaliy Chiley}{equal,cerb}
\mlsysauthor{Abhinav Venigalla}{wdwac}
\mlsysauthor{Joel Hestness}{cerb}
\mlsysauthor{Urs K\"{o}ster}{wdwac,goo}
\end{mlsysauthorlist}

\mlsysaffiliation{cerb}{Cerebras Systems, Inc}
\mlsysaffiliation{tsla}{Tesla}
\mlsysaffiliation{goo}{Google}
\mlsysaffiliation{wdwac}{Work done while at Cerebras}

\mlsyscorrespondingauthor{Atli Kosson}{contact@kosson.is}
\mlsyscorrespondingauthor{Vitaliy Chiley}{vitaliy@cerebras.net}

\mlsyskeywords{Machine Learning, MLSys}

\vskip 0.3in

\begin{abstract}
New hardware can substantially increase the speed and efficiency of deep neural network training. To guide the development of future hardware architectures, it is pertinent to explore the hardware and machine learning properties of alternative training algorithms. In this work we evaluate the use of small batch, fine-grained Pipelined Backpropagation, an asynchronous pipeline parallel training algorithm that has significant hardware advantages. We introduce two methods, Spike Compensation and Linear Weight Prediction, that effectively mitigate the downsides caused by the asynchronicity of Pipelined Backpropagation and outperform existing techniques in our setting. We show that appropriate normalization and small batch sizes can also aid training. With our methods, fine-grained Pipelined Backpropagation using a batch size of one can match the accuracy of SGD for multiple networks trained on CIFAR-10 and ImageNet. Simple scaling rules allow the use of existing hyperparameters for traditional training without additional tuning.
\end{abstract}
]



\printAffiliationsAndNotice{\mlsysEqualContribution} 


\section{Introduction} 
\label{sec:intro}
In recent years, the compute requirements for training state of the art deep neural networks have rapidly increased \cite{amodei2018ai}.
To manage the increased compute requirements, new and efficient hardware architectures are being developed for accelerating deep learning.
Traditional deep learning accelerators rely on batch parallelism (sometimes called data parallelism) to hide memory bandwidth and latency issues.
This can scale to large batches \cite{shallue2019measuring} but can have considerable overheads that limit hardware efficiency (see Appendix~\ref{sec:DPefficiency}).
New hardware architectures could support other, potentially more efficient, training techniques and may not be well suited for traditional methods.
To guide the development of future hardware, it is therefore important to evaluate alternative training techniques and understand their advantages and limitations compared to traditional training.

\begin{figure}[t!]
    \centering
    \includegraphics[width=.975\linewidth]{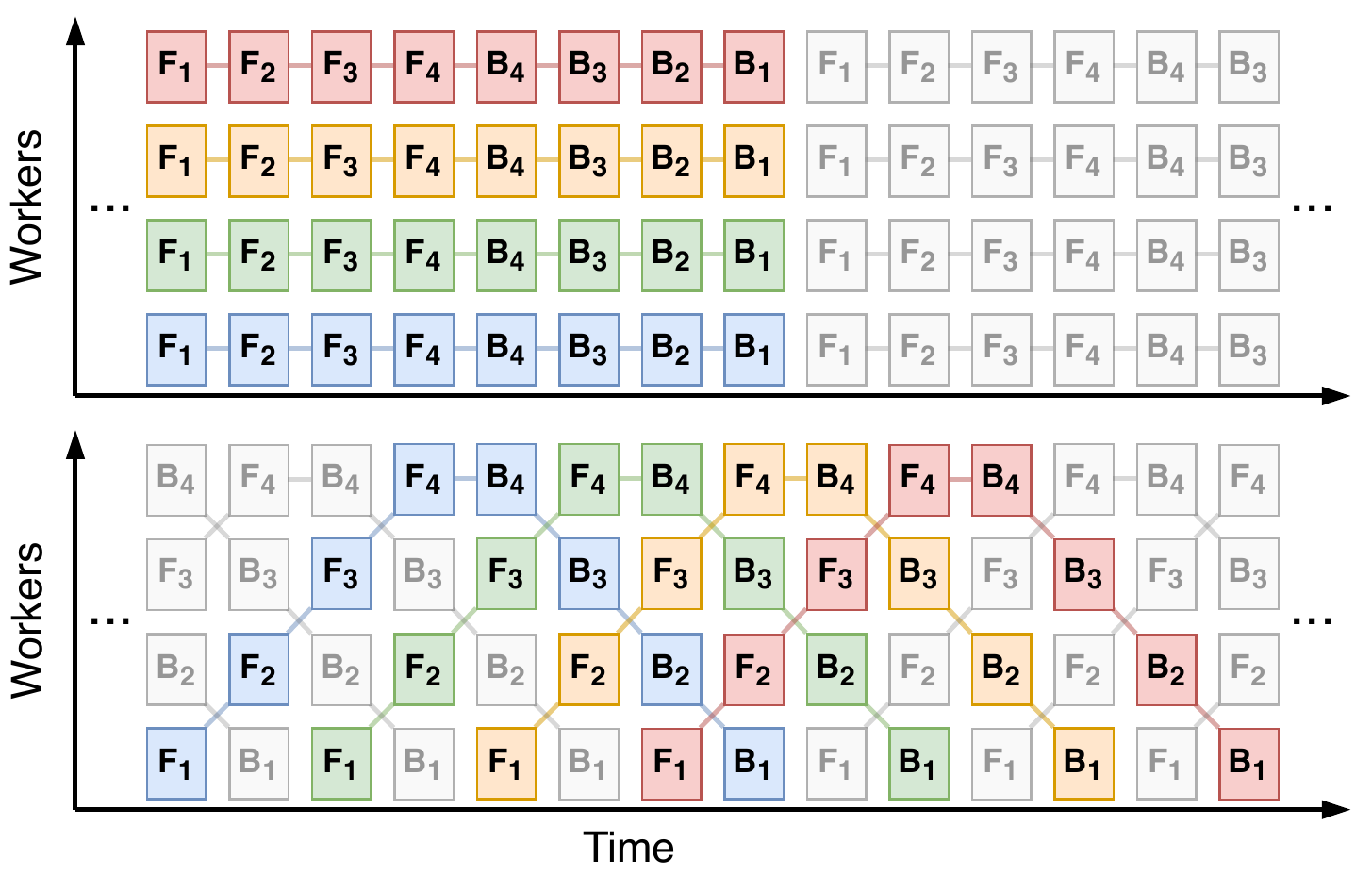}
    \caption{
        \textbf{Top:} batch parallelism. 
        \textbf{Bottom:} pipeline parallelism. 
        We show four workers and a network that can be split into four sequential transformations $F_1, F_2, F_3, F_4$ with corresponding backwards operations $B_1, B_2, B_3, B_4$. 
        The steady state is shown and for simplicity $F$ and $B$ are shown taking the same time. 
        The processing of four inputs is highlighted. 
        In pipeline parallelism workers can specialize to perform a subset of the transformations.
    }
    \label{fig:ppvsdp}
\end{figure}

One alternative to batch parallelism is pipeline parallelism (Figure \ref{fig:ppvsdp}), which divides the model into sequential segments we call pipeline stages.
Each worker is assigned to one stage and inputs proceed sequentially through the stages, similar to an assembly line.
This form of parallelism has the advantage that each worker only performs a subset of the computation which allows them to specialize.

Fine-grained pipeline parallelism assigns one layer of the network to each stage, maximizing opportunities for specialization.
Since each worker only processes one layer it only needs to access a small set of weights which may fit in local memory.
This eliminates bandwidth and latency issues associated with fetching the weights.
For highly configurable architectures, the limited logic set required for each worker allows for an optimized allocation of compute resources between workers, increasing overall utilization.
\citeauthor{zhang2019scalable}~\citeyearpar{zhang2019scalable} find that fine-grained pipelining can enable speedups of up to 3.5x in their setting.
\citeauthor{li2017caterpillar}~\citeyearpar{li2017caterpillar} and \citeauthor{chen2016eyeriss}~\citeyearpar{chen2016eyeriss} both report energy savings of up to 3x.
Fine-grained pipelining can also enable efficient sparse processing which \citeauthor{chen2019eyeriss}~\citeyearpar{chen2019eyeriss} show can result in up to a 42.5x and an 11.3x improvement in throughput and
energy efficiency, respectively.

\begin{figure}
    \centering
    \includegraphics[width=\linewidth]{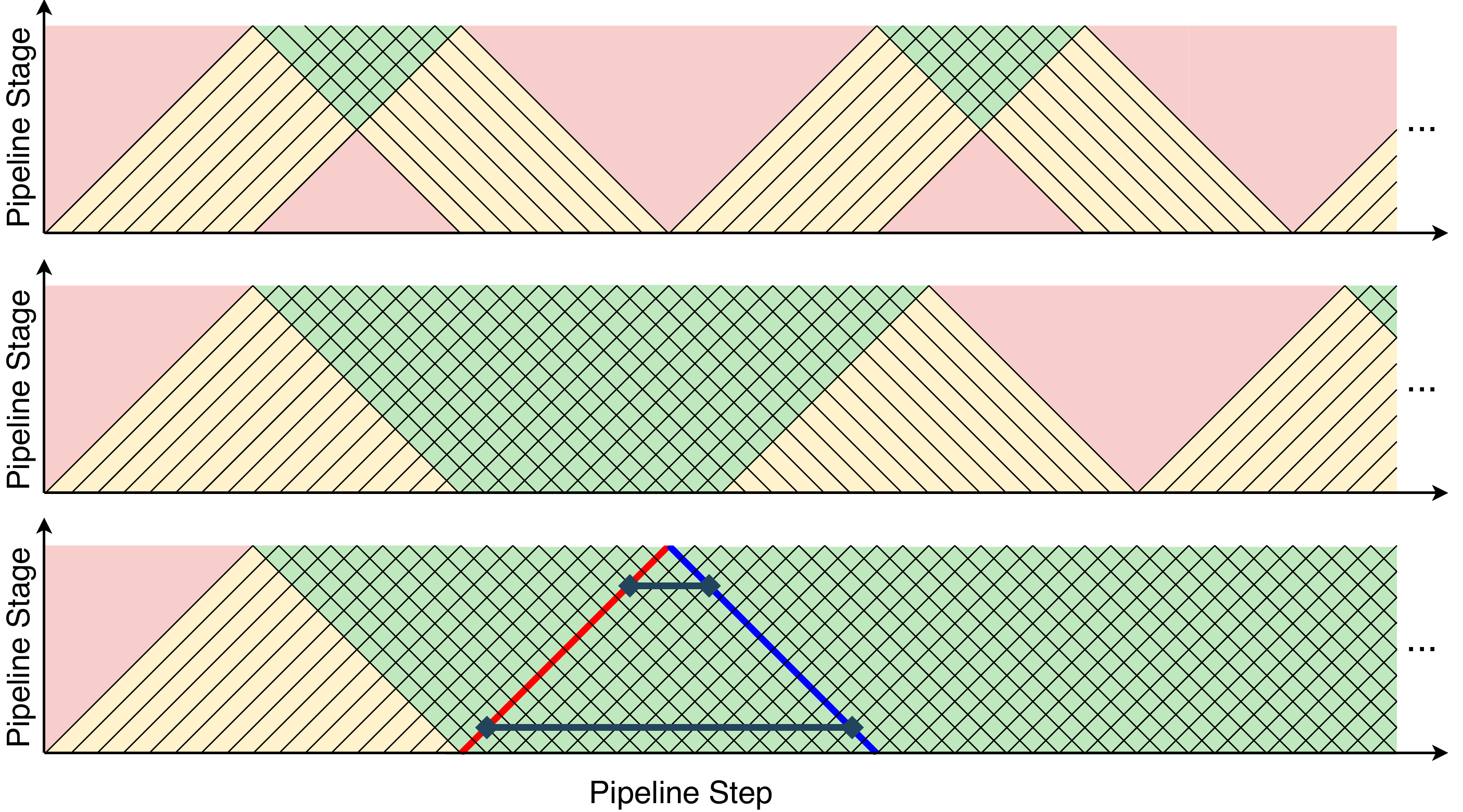}
    \caption{
    Utilization of different pipeline parallel modes. 
    Idle workers are depicted in red, fully utilized workers in green and partially utilized workers (only processing either the forward or backward pass while filling or draining the pipeline) in yellow. 
    \textbf{Top:} Small batch size fill and drain SGD. 
    \textbf{Middle:} Large batch size fill and drain SGD. 
    \textbf{Bottom:} Pipelined Backpropagation.
    The red and blue lines show the forward and backward pass of a single sample. 
    The grey lines show the delays for two of the stages.
    }
    \label{fig:fdvspb}
\end{figure}

Pipeline parallel training commonly implements a form of mini-batch SGD \cite{Huang2018GPipeET}.
This is done by dividing the batch into micro-batches that are sequentially fed into the pipeline (filling it) and waiting for the resulting gradients of all micro-batches (draining the pipeline).
When the pipeline is empty, the parameters are updated and the process is then repeated for the next batch.
Filling and draining the pipeline for each update can significantly lower hardware utilization when the update size is small compared to the number of pipeline stages (Figure~\ref{fig:fdvspb}).
Although it is possible to train with large batch sizes, this makes the hyperparameters harder to tune for a given compute budget \cite{shallue2019measuring} and hyperparameters commonly found in literature are tuned for modest batch sizes.

\textit{Pipelined Backpropagation} (PB)~\cite{Ptrowski1993PerformanceAO} is an asynchronous training technique that avoids fill and drain overhead by updating the weights without draining the pipeline first.
This results in \textit{weight inconsistency}, the use of different weights on the forward and backward passes for a given micro-batch.
The weights used to produce a particular gradient may also have been updated when the gradient is applied, resulting in \textit{stale (or delayed) gradients}.
For these reasons PB resembles Asynchronous SGD~\cite{lian2015asynchronous} and is not equivalent to standard SGD.
Fine-grained pipelining increases the number of pipeline stages and hence increases the weight inconsistency and delay.
For a more detailed discussion of Pipelined Backpropagation and how it gives rise to weight inconsistency and gradient staleness, see Appendix~\ref{apdx:pb}.

The formulation of Pipelined Backpropagation we use throughout this work performs an optimization step after every micro-batch.
Unless otherwise specified, we use a micro-batch size of one and adjust the learning rate and momentum to keep the magnitude of a given weight update proportional to the batch size.
This has several advantages.
First, smaller micro-batches result in smaller weight updates and consequently reduce the effect of weight inconsistency and gradient delays in PB.
Secondly, in pipeline parallelism the activation memory requirements have a quadratic dependence on the number of pipeline stages.
Fine-grained pipeline parallelism with large micro-batches may not fit into memory.
Small batch processing may pose an issue for traditional deep learning accelerators, such as GPUs, which commonly rely on large batch sizes to hide low memory latency and amortize the bandwidth required for loading weights.
New hardware architectures with different memory characteristics may not suffer from these issues, especially when the weights are stored in local memory.
Finally, small batch training has the added benefit of stabilizing training, increasing generalization performance \cite{Masters2018RevisitingSB}, and easing hyperparameter tuning~\cite{li2018visualizing}.

\subsection{Related Works}
GPipe~\cite{Huang2018GPipeET} and Megatron-LM~\cite{shoeybi2019megatron} combine data and pipeline parallel training but incur the fill and drain overhead.
Draining the pipeline is required because of the \textit{locking} nature of SGD \cite{jaderberg2017decoupled}.
Unlocking the forward, backward, and update operations is an active area of research~\cite{jaderberg2017decoupled,huo2018training,huo2018decoupled,xu2019diversely,belilovsky2019decoupled,gaunt2017ampnet}.
\citeauthor{Ptrowski1993PerformanceAO}~\citeyearpar{Ptrowski1993PerformanceAO} propose Pipelined Backpropagation which updates the weights without draining the pipeline to avoid the fill and drain overhead but introduces weight inconsistency and stale gradients.

Recent works have also explored mitigating these issues when training networks through a combination of data parallelism and PB \cite{chen2012pipelined,Harlap2018PipeDreamFA,Chen2018EfficientAR,yang2019pipemare,Zhuang2019FullyDN}.
PipeDream~\cite{Harlap2018PipeDreamFA} proposes weight stashing (WS) and vertical sync (VS).
Weight stashing saves the weights used on the forward pass for use on the backwards pass and vertical sync updates the weights at the same moment in time.
\citeauthor{Chen2018EfficientAR}~\citeyearpar{Chen2018EfficientAR} show how weight stashing is ineffective in their setting and propose a form of weight prediction called SpecTrain to mitigate the effects of both stale gradients and inconsistent weights.
PipeMare~\cite{yang2019pipemare} applies discrepancy correction (a form of backward weight prediction) to address inconsistent weights and learning rate rescheduling (a new form of learning rate warmup) to help with stale gradients.
\citeauthor{Zhuang2019FullyDN}~\citeyearpar{Zhuang2019FullyDN} propose Gradient Shrinking, which exponentially decays the gradients for each stage based on the delay. 

Unlike prior work, we completely replace batch parallelism with fine-grained pipelined parallelism at batch size one.
Compared to previous works this increases the number of pipeline stages and allows for greater worker specialization.
We find that in our setting existing mitigation methods such as Weight Stashing, Gradient Shrinking
and SpecTrain are insufficient while some others, such as Features Replay~\cite{huo2018training}, do not apply since each stage has a single layer.
Our contributions are as follows:
\begin{itemize}[noitemsep,topsep=0pt]
    \item We explore the hardware aspects of small batch size, fine-grained, Pipelined Backpropagation for training deep neural networks and how Coarse-Grained Reconfigurable Arrays~\cite{podobas2020survey} can be particularly well suited for this sort of training.
    \item We propose two methods, \textit{Spike Compensation} and \textit{Linear Weight Prediction} to mitigate the drawbacks of PB: inconsistent weights and stale gradients. An analysis on a simplified model shows how they can counteract the effects of stale gradients and restore the benefits of momentum in the presence of delays. We show that our methods outperform existing techniques in the fine-grained small batch PB setting, without additional hyperparameters to tune.
    \item We show the importance of small batches and proper normalization for this type of training. Simple scaling rules enable the use of existing hyperparameters for traditional training without further tuning. With our methods, this makes fine-grained PB a drop-in replacement for mini-batch SGD training on standard image classification benchmarks, CIFAR-10~\cite{c10bib} and ImageNet~\cite{imagenet_cvpr09}.
\end{itemize}


\section{Fine Grained Pipeline Parallelism on Hardware}

\label{sec:fgpponHW}
\begin{figure}
    \centering
    \begin{minipage}[t]{0.45\linewidth}
        \centering
        \includegraphics[width=\linewidth]{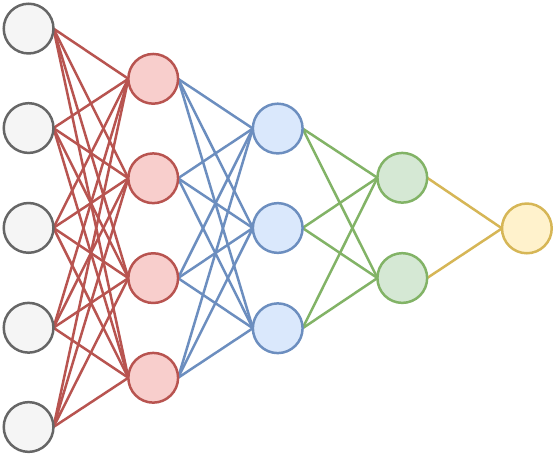}
        \label{fig:nnonfab_nn}
    \end{minipage}
    \hskip .1in
    \begin{minipage}[t]{0.387\linewidth}
        \centering
        \includegraphics[width=\linewidth]{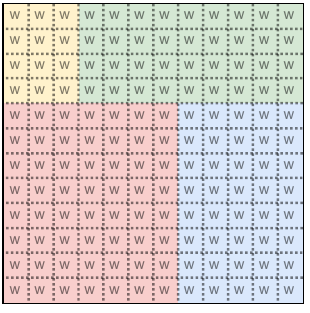}
        \label{fig:nnonfab_fab}
    \end{minipage}
    \caption{
        Neural network (Left) computation spatially distributed across a compute fabric (Right) with many workers (w). Each worker is  specialized and an appropriate amount of compute (workers) is allotted to each layer of the NN. The pipeline depth is defined by the network topology and not the compute architecture.
    }
    \label{fig:nnonfab}
\end{figure}

Fine grained pipeline parallel training can allow workers to specialize, increasing utilization and accelerating neural network training.
A Coarse-Grained Reconfigurable Array (CGRA) is an example of a hardware architecture that can significantly benefit from this.
CGRAs are a grid of locally connected cores with fast interconnects and a distributed memory architecture.
Pipeline parallelism can be performed by spatially distributing the layers over the compute array (Figure~\ref{fig:nnonfab}).
Due to its distributed nature, the overall on-chip memory in a CGRA can be significantly larger than on traditional architectures, allowing workers to store weights locally.
This enables persistent kernel execution eliminating any kernel launch overheads as well as overcoming bandwidth and latency issues for small batch processing.
Using a small micro-batch size limits the memory required for storing activations limiting the overall memory required.

Standard \textit{layer sequential}, large batch size training can be performed on CGRAs, but may require adding off-chip memory.
Layer sequential execution eliminates the benefits of persistent kernel execution and worker specialization.
Distributing model weights and reducing gradients across to all cores creates latency issues similar to those encountered in distributed training that can be difficult to mask.
Fine-grained, small micro-batch size pipelining may therefore be the best way to achieve high utilization levels and has been shown to increase processing speed and energy efficiency.
Architectures with distributed low latency memory architectures, such as \cite{lie2020wafer,vassilieva2020neural,sambanova2020whitepaper,nicol2017coarse,jia2019dissecting}, could benefit from pipelined training approaches.
\citeauthor{sambanova2020whitepaper}~\citeyearpar{sambanova2020whitepaper} shows how this spatial distribution of layers can be used to accelerate neural networks on their Reconfigurable Dataflow Architecture.
GraphCore presents how pipelined training improves the IPU's efficiency \cite{graphcore2020doc}, and 
\citeauthor{lie2020wafer}~\citeyearpar{lie2020wafer} and \citeauthor{vassilieva2020neural}~\citeyearpar{vassilieva2020neural}
discuss how fine-grained pipelined parallelism can accelerate neural network training on the Cerebras Wafer-Scale Engine.

\subsection{Simulating Pipelined Parallelism on GPUs}
\label{sec:gprop}

Since suitable hardware architectures are not widely available yet, we are not able to test our methods in a hardware setup where we expect speedups.
GPUs are designed for layer sequential batch processing and are not efficient for pipeline parallel training with small batch sizes. 
However, they are readily available so we opt to use them for this exploratory work.
In this section we describe how we simulate pipeline parallel training on GPUs.
Appendix~\ref{sec:gpu_pp_lim} discuses the GPU hardware limitations that prevent this mode of training from being efficient on GPUs.
Therefore, our goal is only to make our experiments feasible on GPUs, not to speed up training compared to well tuned batch parallelism.

\begin{figure}
    \centering
    \includegraphics[width=\linewidth]{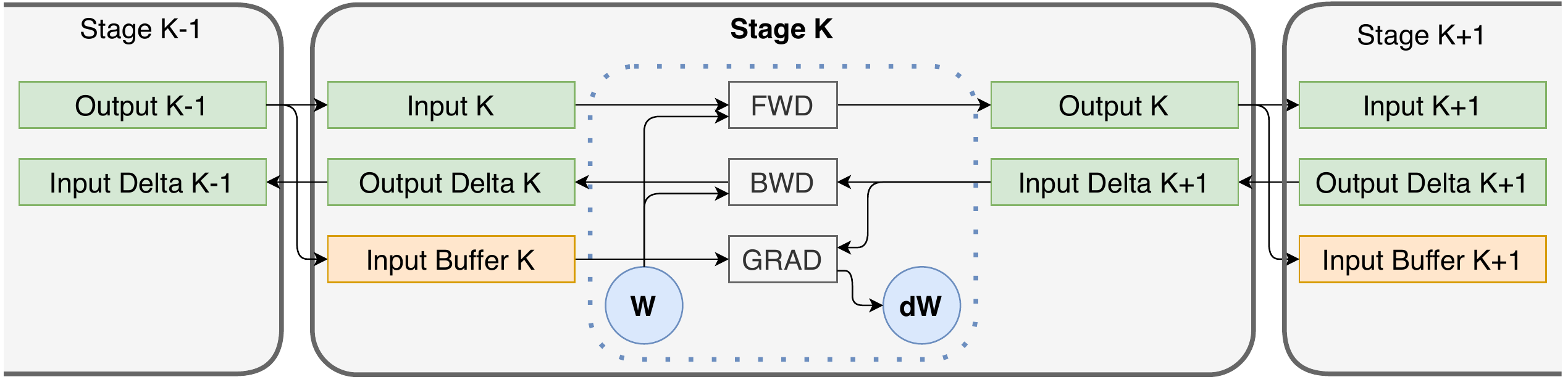}
    \caption{Pipeline Stage. The network is divided into stages that correspond to pipeline stages. Each stage contains logic for the forward, backward, and gradient computation for the corresponding network component. The use of buffers enables parallel execution, one stage can compute a new output while another stage is using the previous output.}
    \label{fig:gprop_stage} 
\end{figure}

In particular we are interested in simulating fully pipeline parallel training on networks such as ResNet-50 with a maximal number of pipeline stages and no batch parallelism.
Most modern deep learning frameworks are not well suited for such experimentation.
To enable efficient simulation of fully pipeline parallel training, we built a mini-framework implemented in C++ using cuDNN~\cite{chetlur2014cudnn} kernels, custom CUDA kernels, and Thrust.

The network is split into structures we call stages that act as pipeline stages. 
Each stage manages all resources needed to compute the forward and backward passes for the corresponding part of the network (Figure~\ref{fig:gprop_stage}). 
The forward, backward, and gradient computations within a stage are completely unlocked and run in parallel.
In our experiments we sometimes group several network components together into a single stage. 
One example of this is grouping convolution, normalization, and ReLU into one stage.
We use CUDA streams to run the stages in parallel. 
The framework also supports splitting the network over multiple GPUs and uses a different thread to launch the stages on each GPU.


\section{Methods}
\label{sec:methods}

We introduce two compensation methods for Pipeline Backpropagation: Linear Weight Prediction and Spike Compensation.
We formulate them for SGD with momentum\footnote{Both methods require momentum and can be adapted for other momentum based optimizers.} 
(SGDM) which we write as:
\begin{align}
    v_{t+1} &= m v_t + g_t \label{eqn:sgdmv}\\
    w_{t+1} &= w_t - \eta v_{t+1} 
    \label{eqn:sgdm}
\end{align}
where $w_t$ are weights (parameters) at time $t$, $v_t$ is the velocity (sometimes called momentum), $m$ is the momentum coefficient, $\eta$ is the learning rate, and
$g_t$ represents the gradient estimate applied at time $t$.
This estimate may be delayed, and could be calculated with inconsistent weights.

PB partitions the network into $S$ stages.
Each stage has its own delay, $D^s$ for $s\in [0,...,S-1]$, determined by the network architecture.
We describe and analyze our methods for a constant delay, $D$, without modeling the pipeline or inconsistency. 
When we use the methods for PB we apply them to each stage separately, with the corresponding delay set to the number of steps between the forward and backwards passes for that stage.
To simplify notation we drop the superscript $s$ representing the stage index.
We write the gradient as a function of the weights alone, whereas in SGD the gradient may also depend on inputs or other data.


\subsection{Spike Compensation}

\begin{figure}
    \centering
    \includegraphics[width=\linewidth]{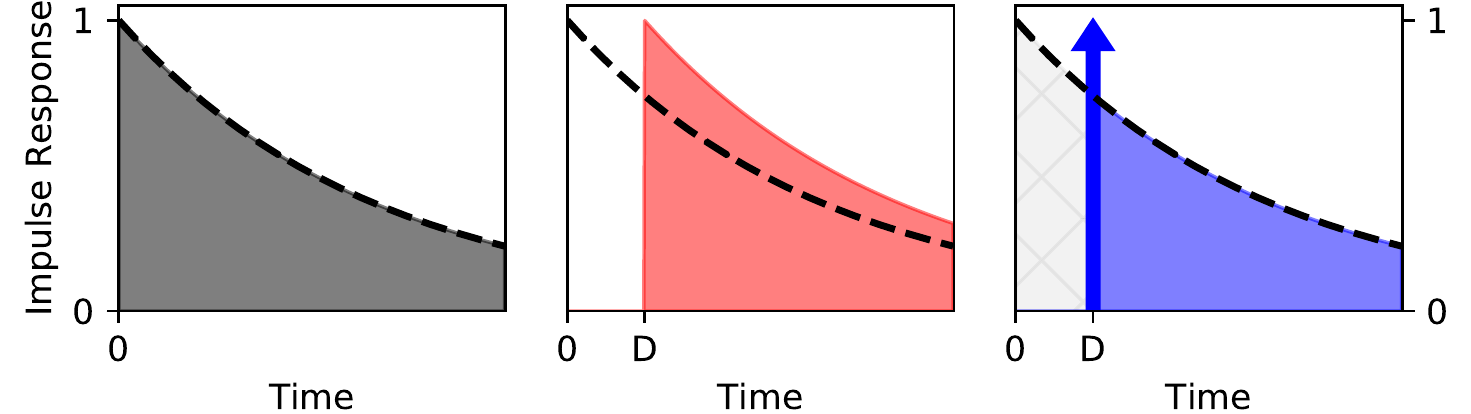}
    \caption{
    \textbf{Left:} Momentum exponentially smooths gradients over time so the contribution of each gradient to future weights updates (the impulse response) is an exponentially decaying function from the time it arrives.
    \textbf{Middle:} A delayed gradient has an impulse response shifted by the delay $D$. The dotted line shows the baseline without delay.
    \textbf{Right:} With spike compensation (SC\textsubscript{D}) the impulse response has a spike (denoted with an arrow) and then matches the no-delay case. The size of the spike matches that of the missed updates compared to the baseline shown in light gray.
    }
    \label{fig:sc}
\end{figure}

We introduce Spike Compensation (SC) to mitigate the effects of delayed gradients in PB. The method uses a modified weight update which increases the contribution of the latest gradient relative to the velocity. For a delay of $D$ this can generally be written as:
\begin{align}
    & g_t = G(w_{t-D}) \label{eq:gscg} \\
    & v_{t+1} = m v_t + g_t \label{eq:gscv} \\
    & w_{t+1} = w_t - \eta \cdot \left( a v_{t+1} + b g_t \right) \label{eq:gscw}
\end{align}
where $a$ and $b$ are functions of the delay. We could absorb either $a$ or $b$ into $\eta$ but use this form to keep $\eta$ consistent with other methods. We refer to this form as generalized Spike Compensation (GSC). To reason about sensible choices for $a$ and $b$ we can look at the contribution of each gradient over time in the no-delay case vs the delay case (see Figure~\ref{fig:sc}). When a gradient $g$ is obtained with some delay $D$, this gradient would already have contributed to $D$ weight updates in the no-delay case. 
The total contribution of the gradient so far would have been:
\begin{equation}
    \sum_{t=0}^{D-1} m^t g = \frac{1-m^D}{1-m} g
\end{equation}

This inspires our default choice of $a$ and $b$ for Spike Compensation which we will refer to as SC\textsubscript{D}:
\begin{align}
    a = m^D \quad \text{and} \quad b = \frac{1-m^D}{1-m} \label{eq:scd} 
\end{align}
For this choice, the missing weight update is applied immediately and the contribution of the gradient at later time steps will match that of the no-delay case. The total contribution of each gradient to the weights over the course of training is unchanged, this only changes how the gradients are applied over time. 
The modified weight update can equivalently be seen as approximating the velocity in the no-delay case with $a v_{t+1} + b g_t$. 
This uses the latest gradient to approximate the gradient terms in the velocity that have not been observed due to the delay.
For a delay of zero, SC\textsubscript{D} reduces to SGDM~\eqnref{eqn:sgdm}.


\subsection{Linear Weight Prediction}
Both the weight inconsistency and gradient delay arise from the fact that we can not access the (future) weights used on the backwards pass when we compute the forward pass. The goal of weight prediction is to approximate the backwards weights on the forward pass. 
We want to approximate:
\begin{equation}
    w_{t+D} = w_t - \eta \sum_{k=0}^{D-1} v_{t+k+1}
\end{equation}
where $D$ is the delay (number of update steps between the forward and backwards passes). The future velocities are unknown but can be approximated by assuming a constant gradient $\hat{g}$ over the prediction horizon, i.e. the number of iterations over which the prediction is made. This gives:
\begin{equation}
    v_{t+k+1}\!\approx\!m^k v_{t+1} + \hat{g}\!\sum_{i=0}^{k-1} m^i = m^k v_{t+1} + \frac{1-m^k}{1-m} \hat{g}
\end{equation}
which results in predicted weights:
\begin{align}
    \hat{w}_{t+D}\!=\!w_t\!-\!\eta\frac{1\!-\!m^D}{1\!-\!m}v_{t+1}\!-\frac{\eta \hat{g}}{1\!-\!m}\!\left(\!D\!-\!\frac{1\!-\!m^D}{1\!-\!m}\!\right)
    \label{eq:wpvariants}
\end{align}
We have several good choices $\hat{g}$ including setting it to zero or estimating it based on recent gradients. In this work we focus on weight prediction where the direction of the velocity does not change, i.e. $\hat{g}$ is collinear with $v_t$. We refer to this as linear weight prediction (LWP). The approximation for the weights at time $t$ and delay $D$ can then be written in terms of past weights and velocities as:
\begin{align}
    \hat{w}(t,D,T) = w_{t-D} - \eta T v_{t-D} =: \hat{w}_{\text{v}}(t,D,T) \label{eq:lwp_v_form}
\end{align}
Where $T$ is a hyperparameter we call the horizon of the weight prediction. For SGDM without modifications, we can equivalently write the approximate in terms of the previous weights alone:
\begin{align}
    \hat{w}(t,D,T) & = w_{t-D} + T \cdot (w_{t-D} - w_{t-D-1}) \nonumber \\  & =: \hat{w}_{\text{w}}(t,D,T) \label{eq:lwp_w_form}
\end{align}
When combined with Spike Compensation, or potentially when using other optimizers, the predictions given by equations \eqnref{eq:lwp_v_form} and \eqnref{eq:lwp_w_form} differ. When this is the case we refer to the two types as LWP$^\text{v}$ (velocity form) and LWP$^\text{w}$ (weight difference form), respectively. The update step is:
\begin{align}
    & g_t = G\left(\hat{w}(t,D,T) \right) \label{eq:lwpg} \\
    & v_{t+1} = m v_{t} + g_t \label{eq:lwpv} \\
    & w_{t+1} = w_t - \eta v_{t+1} \label{eq:lwpw}
\end{align}
In the rest of this paper we use LWP\textsubscript{D} to denote LWP with our default choice of $T=D$. This is equivalent to choosing $\hat{g}=(1-m) v_{t+1}$ in \eqnref{eq:wpvariants} which would result in a constant velocity. This form is closely related to the weight prediction used in SpecTrain~\cite{Chen2018EfficientAR} which extends the prediction horizon and also predicts weights on the backwards pass (see Appendix~\ref{apdx:wp_forms}).

\begin{figure*}
    \centering
    \includegraphics[width=\linewidth]{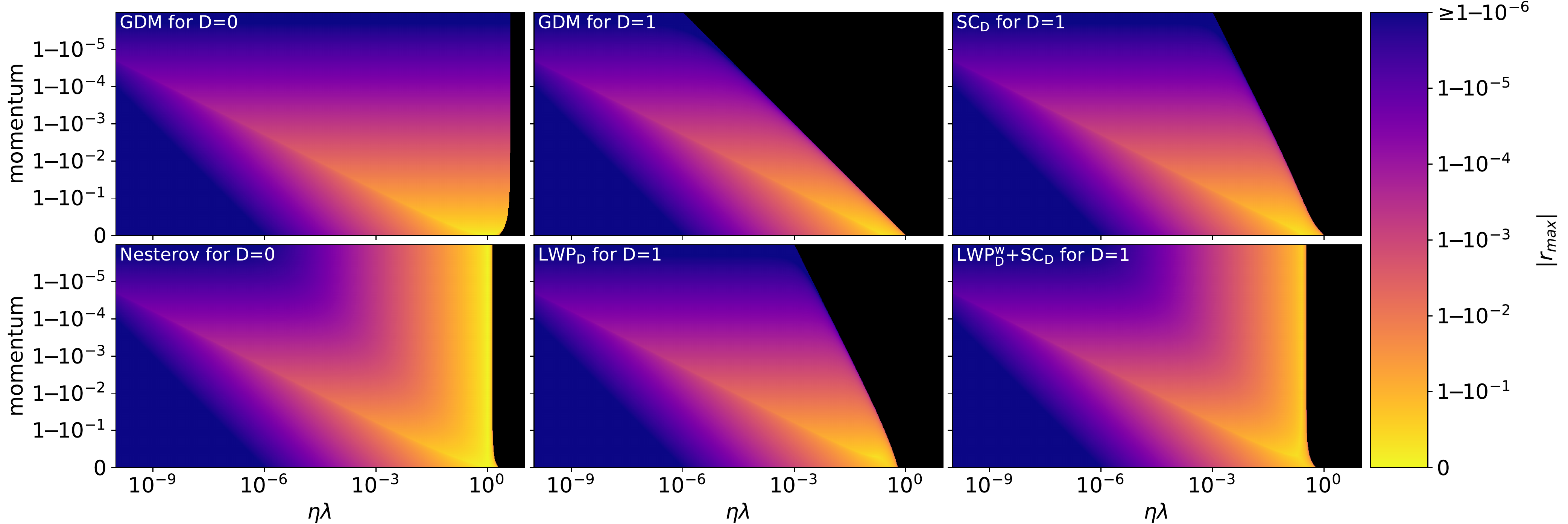}
    \caption{These plots show the magnitude of the dominant root of the characteristic polynomials given in equations \eqnref{eq:char_sgdm}-\eqnref{eq:char_cmb} as a function of the normalized rate $\eta\lambda$ and the momentum $m$. The two leftmost plots show zero delay baselines and the other plots use a delay of $D=1$. The blacked out region has roots with magnitudes larger than one and is therefore unstable. For a delay of one, Nesterov momentum is equivalent to Spike Compensation, but for larger delays this does not hold and Nesterov is only marginally better than GDM.}
    \label{fig:RootHeatmaps}
\end{figure*}


\subsection{Combined Mitigation}
\label{sec:combined_mitigation}

Spike Compensation and weight prediction can be combined resulting in the following update step:
\begin{align}
    & g_t = G\left( \hat{w}(t,D,T) \right) \label{eq:wpscg}\\
    & v_{t+1} = m v_{t} + g_t \label{eq:wpscv} \\
    & w_{t+1} = w_t - \eta \cdot \left( a v_{t+1} + b g_t \right) \label{eq:wpscw}
\end{align}
where, as before, $T$ is the horizon of the weight prediction and $a$ and $b$ are the coefficients for the Spike Compensation. When combined with Spike Compensation $\hat{w}_{\text{v}}(t,D,T) \ne \hat{w}_{\text{w}}(t,D,T)$.
In the combination $\hat{w}_{\text{w}}(t,D,T)$ can be interpreted as using Spike Compensation to approximate the velocity used in the weight prediction which also corresponds to a different choice of $\hat{g}$ in equation \eqnref{eq:wpvariants} using the most recent gradient estimate.


\subsection{Analysis for a Convex Quadratic}
\label{sec:cq_analysis}
In this section we analyze the optimization of a convex quadratic loss with gradient delay. We find that delay causes effects that can generally not be mitigated through hyperparameter tuning but our methods: 
\begin{itemize}[noitemsep,topsep=0pt]
    \item Improve convergence for large condition numbers
    \item Allow higher learning rates for large momentum values
    \item Restore the benefits of momentum for poorly conditioned losses
\end{itemize}
Although the loss surfaces of convex quadratics are much simpler then those of neural networks, they can be a useful tool to gain a high level understanding of various aspects of neural network optimization.
For instance, \citeauthor{zhang2019algorithmic}~\citeyearpar{zhang2019algorithmic} accurately model the effect of batch size on neural network optimization through a noisy quadratic model.
\citeauthor{venigalla2020adaptive}~\citeyearpar{venigalla2020adaptive}, \citeauthor{giladi2019stability}~\citeyearpar{giladi2019stability}, and \citeauthor{yang2019pipemare}~\citeyearpar{yang2019pipemare} use quadratic models to analyze gradient delay mitigation methods for use in asynchronous deep network training.
In general we find that the insights from the convex quadratic hold surprisingly well for neural networks as shown in our experiments and the appendix.

We follow a similar approach as \cite{o2015adaptive,goh2017why} and write the loss in terms of an eigenbasis of the quadratic as:
\begin{align}
    & L(\phi) = \phi^T \Lambda \phi, & \Lambda = \text{diag}(\lambda^{(1)}, ..., \lambda^{(N)})
\end{align}
where $\phi = [\phi^{(1)}, ... ,\phi^{(N)}]^T$ correspond to the parameters being optimized and $\lambda^{(1)} \ge ... \ge \lambda^{(N)} > 0$ are the eigenvalues of the quadratic. As shown in e.g. \citeauthor{goh2017why}~\yrcite{goh2017why}, any positive definite quadratic can be written in this form through a coordinate transformation. Since $\Lambda$ is diagonal, each coordinate of the gradient $\nabla_{\phi} L(\phi) = \Lambda \phi$ is independent of other coordinates. This allows us to analyze the convergence for each coordinate separately. For simplicity we assume that the gradient is deterministic. A similar analysis would hold for the expected values of $\phi$ if each gradient sample was assumed to be noisy but unbiased. 

In Appendix~\ref{apdx:state_transition} we derive the state transition equations for SGDM with delay and our methods. Since the gradient here is linear, and the coordinates are independent, inserting it into the transition equations results in a linear recurrence relation for each coordinate.  For component $\phi^{(k)}$, with associated eigenvalue $\lambda=\lambda^{(k)}$, the characteristic polynomial for the recurrence relation of each method is:
\begin{align}
    \text{GDM:}\; p(z) = &\; z^{\tau+1}-(1+m)z^\tau+m z^{\tau-1} - \eta \lambda \label{eq:char_sgdm}\\
    \text{GSC:}\; p(z) = &\; z^{\tau+2} - (1+m)z^{\tau+1} + m z^{\tau} \nonumber \\ 
        & + \eta \lambda \cdot (a+b)z - \eta\lambda m b \label{eq:char_gsc}
\end{align}
\begin{align}
    \text{LWP:}\; p(z) = &\; z^{\tau+2} - (1+m)z^{\tau+1} + m z^{\tau} \nonumber \\ 
        & + \eta \lambda \cdot (1+T)z - \eta\lambda T \label{eq:char_lwp} \\
    \text{LWP}^\text{w}\!\!+\!\text{GSC:} \nonumber \\
    p(z) = &\; z^{\tau+3} - (1+m)z^{\tau+2} + m z^{\tau+1}  \nonumber \\ 
    & + \eta\lambda\cdot (a+b)(T+1) z^{2} \nonumber \\
    & - \eta\lambda\cdot \left((T+1)mb + T\cdot(a+b)\right)z \nonumber \\
    & + \eta\lambda T m b \label{eq:char_cmb}
\end{align}
where GDM stands for gradient descent with momentum, GSC is general Spike Compensation, LWP is linear weight prediction, $z$ parameterizes the polynomials and other symbols have the same meaning as in Section~\ref{sec:combined_mitigation}. Note that since the gradient is linear, GSC and LWP are equivalent for a certain choice of $a$, $b$ and $T$ as shown in Appendix~\ref{apdx:state_transition}.
Even though this is the case, the characteristic polynomial of the combination cannot be obtained from either method.

Linear recurrence relations have a well known solution in terms of the roots of the corresponding characteristic equation. The resulting sequence for component $\phi^{(i)}$, corresponding to the characteristic polynomial $p(z)$ with roots $r_1,... ,r_n$, can be written as:
\begin{equation}
    \phi_t^{(i)} = \sum_{k=1}^{n} q_k(t) r_k^t \label{eq:xtclosed}
\end{equation}
where $q_k(t)$ is a polynomial. The order of the polynomial is one less than the multiplicity of the corresponding root $r_k$. The coefficients of the polynomials are determined by the initial conditions. 

For our analysis we assume that all components start with some error and look at the rate of convergence in the limit $t \rightarrow \infty$. A component $\phi^{(i)}$ converges to the optimal value of $0$ if $|r_{\mathrm{max}}|=\max_{k}(|r_k|) < 1$. In the limit, the slowest term of equation~\eqnref{eq:xtclosed} will dominate so the error for this component, $\varepsilon^{(i)}$ will be:
\begin{equation}
    \varepsilon^{(i)}_t = |\phi_t^{(i)}-0| \propto |r_{\mathrm{max}}|^t
\end{equation}

\begin{figure}
    \centering
    \includegraphics[width=\linewidth]{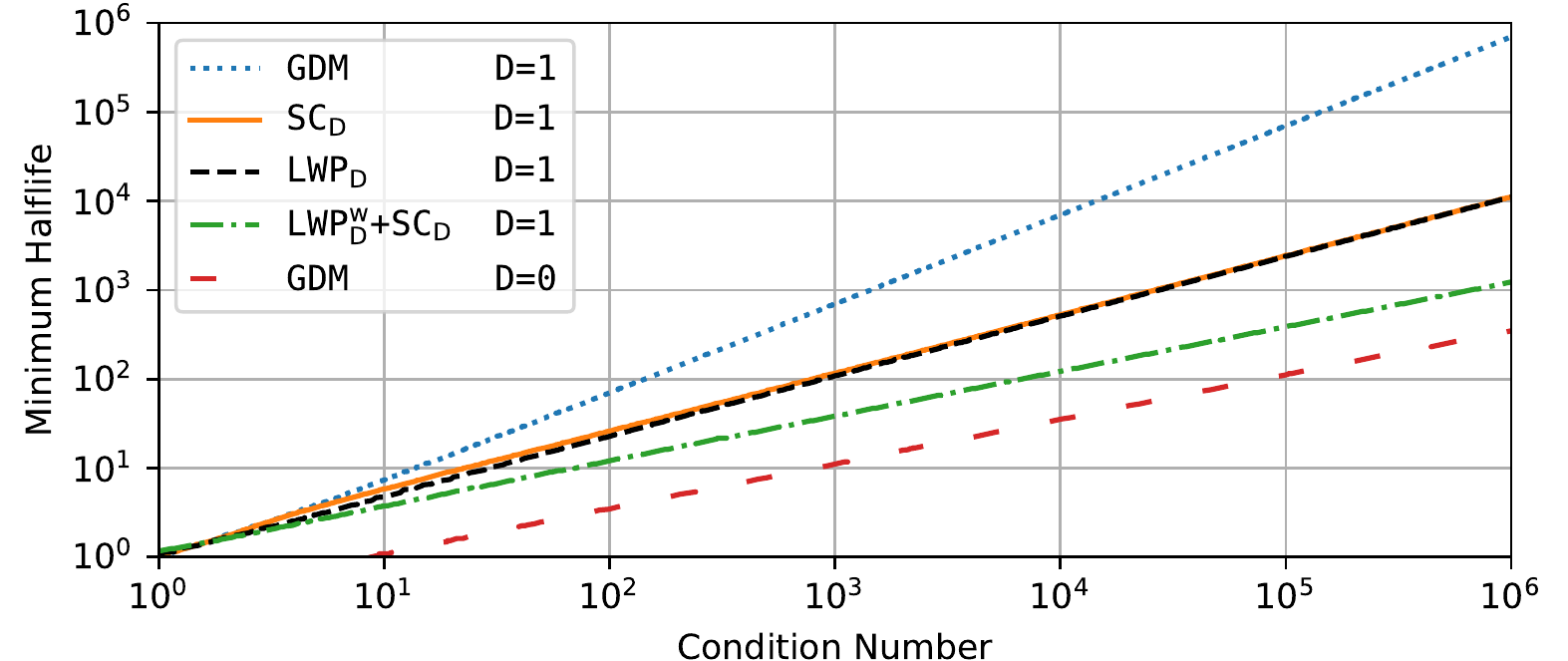}
    \caption{The half-life of the error as a function of the conditioning number when optimizing a convex quadratic with delay $D$. All methods improve the convergence rate, LWP$\,_\text{D}^\text{w}$+SC\textsubscript{D} performs best.}
    \label{fig:HalflifeVsKappa}
\end{figure}

The overall rate of convergence is determined by the slowest component. 
The slowest component depends on the roots of the characteristic polynomial. These can be difficult to determine analytically, so we turn to computational analysis.
For a given delay, we can compute the roots of the characteristic polynomials~\eqnref{eq:char_sgdm}-\eqnref{eq:char_cmb}, including $|r_{\mathrm{max}}|$, as a function of the normalized rate $\lambda \eta$ and the momentum $m$. Figure~\ref{fig:RootHeatmaps} shows heatmaps of $|r_{\mathrm{max}}|$ for each method for a delay of one and our default values of $a$, $b$ and $T$. Note that the region of stability is significantly reduced by the delay, especially for large momentum values. Our compensation methods counteract this, allowing larger learning rates to be used for high momentum values. SC\textsubscript{D} in particular strictly increases the region of stability, the other methods slightly decrease it for small momentum coefficients. 

Figure~\ref{fig:RootHeatmaps} also allow us to reason about more than a single component at a time. Let's assume that we have multiple components, a condition number $\kappa = \lambda_1/\lambda_N$ and a dense spectrum of eigenvalues between $\lambda_1$ and $\lambda_N$. The same learning rate $\eta$ and momentum $m$ are used for all components. The overall convergence rate is determined by the component with the largest $|r_{\mathrm{max}}|$. This corresponds to the largest value in a horizontal line segment between $\eta\lambda_N$ and $\eta\lambda_1$ on the root heatmaps. With a log scale the line segment has a constant length determined by $\kappa$.

\begin{figure}
    \centering
    \includegraphics[width=\linewidth]{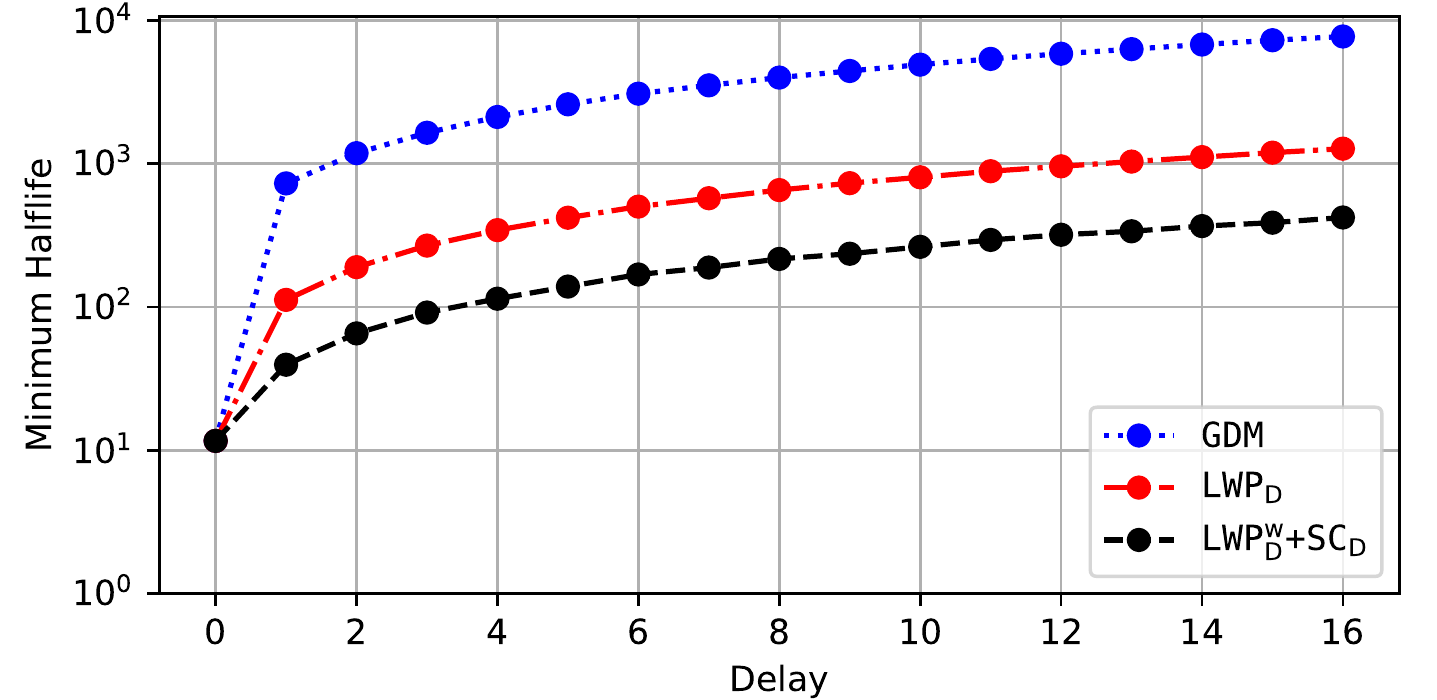}
    \caption{The optimal half-life of the error for different delays when optimizing a convex quadratic with $\kappa=10^3$.}
    \label{fig:HalflifeVsDelay}
\end{figure}

Figure~\ref{fig:HalflifeVsKappa} shows the convergence speed as a function of $\kappa$ for the different methods. We measure the half-life $-\ln{2}/\ln{|r^*|}$ where $|r^*|$ is obtained by finding the lowest max magnitude over all intervals of sufficient length. The methods improve the rate of convergence compared to the delayed baseline. The combination performs the best which also holds for larger delays as is shown in Figure~\ref{fig:HalflifeVsDelay}.

\begin{figure}
    \centering
    \includegraphics[width=\linewidth]{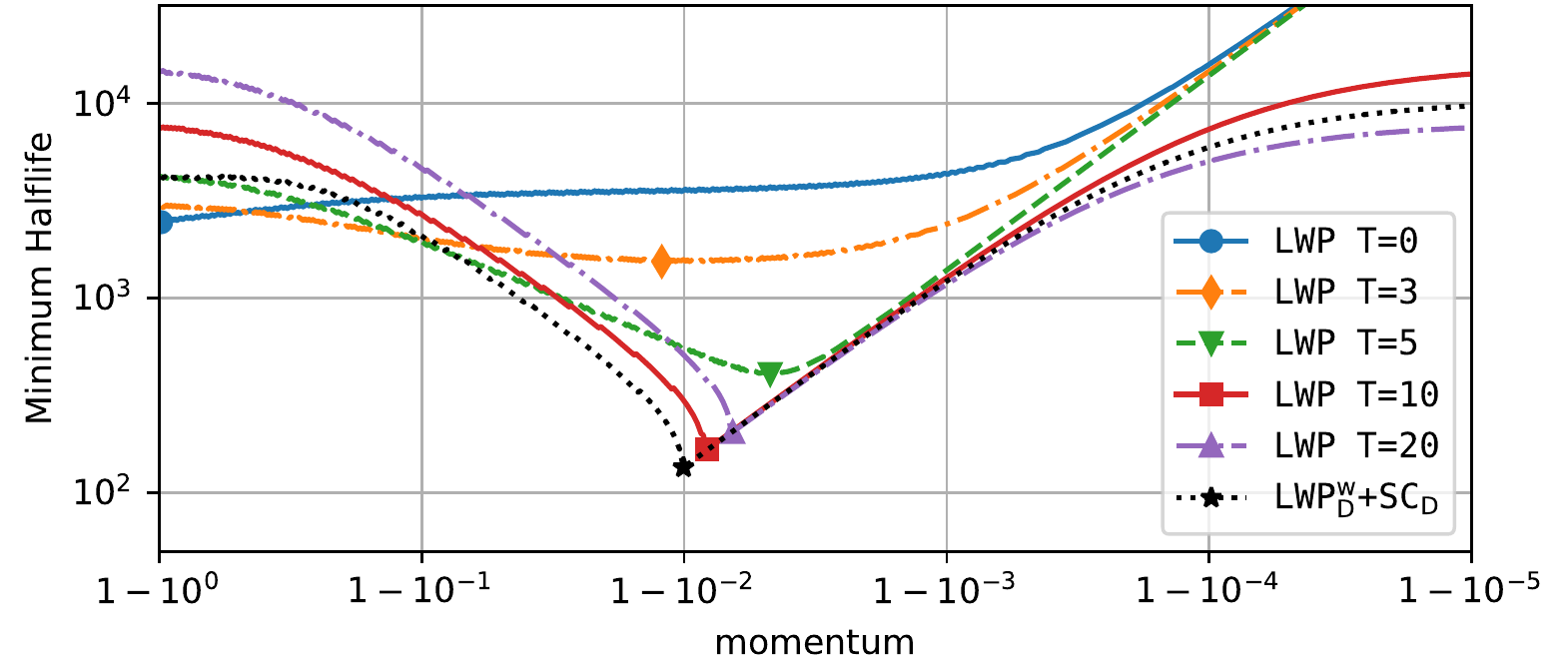}
    \caption{The effect of momentum and the horizon $T$ for weight prediction on the optimal half-life when optimizing a convex quadratic with $\kappa=10^3$ for a delay $D=5$.}
    \label{fig:HalflifeVsMomentumHorizons}
\end{figure}

As mentioned earlier, GSC and LWP can be equivalent for a convex quadratic. The fact that LWP\textsubscript{D} slightly outperforms SC\textsubscript{D} indicates that our selection of $T=D$ is better than the selection of $a$ and $b$ as given in equation~\eqnref{eq:scd} in this case.
Figure~\ref{fig:HalflifeVsMomentumHorizons} shows the effect of different values of $T$. It shows that values close to $T=2D$ are optimal but do not outperform the combination LWP$\,_\text{D}^\text{w}$+SC\textsubscript{D}.
This seems to indicate that ``overcompensating" for the delays, by predicting weights further out in LWP or equivalently by using larger spikes in SC, seems to produce better optimization trajectories. The resulting root heatmaps resemble the ones for the no-delay Nesterov baseline (see LWP$\,_\text{D}^\text{w}$+SC\textsubscript{D} in Figure~\ref{fig:RootHeatmaps}, LWP with $T=2D$ looks similar). Note that adding Nesterov to the delay is not sufficient to get this effect. In Appendix~\ref{apdx:wp2} we show the effect of extended horizons for both the convex quadratic and a neural network.

Figure~\ref{fig:HalflifeVsMomentumHorizons} also reveals that without mitigation ($T=0$ is equal to GDM with delay), the optimal momentum is zero. In the no-delay case the optimal momentum is given by $m=\left((\sqrt{\kappa}-1)/(\sqrt{\kappa}+1)\right)^2$ \cite{zhang2017yellowfin} which increases with the condition number. Our compensation methods restore the benefits of momentum for high condition numbers. Overall the combined mitigation performs the best. Extended horizons for LWP or the equivalent coefficients for GSC also outperform our default choice in this case but are unable to match the combination LWP$\,_\text{D}^\text{w}$+SC\textsubscript{D}.

\begin{figure*}
    \begin{minipage}[b]{.48\textwidth}
        \centering
        \includegraphics[width=\linewidth]{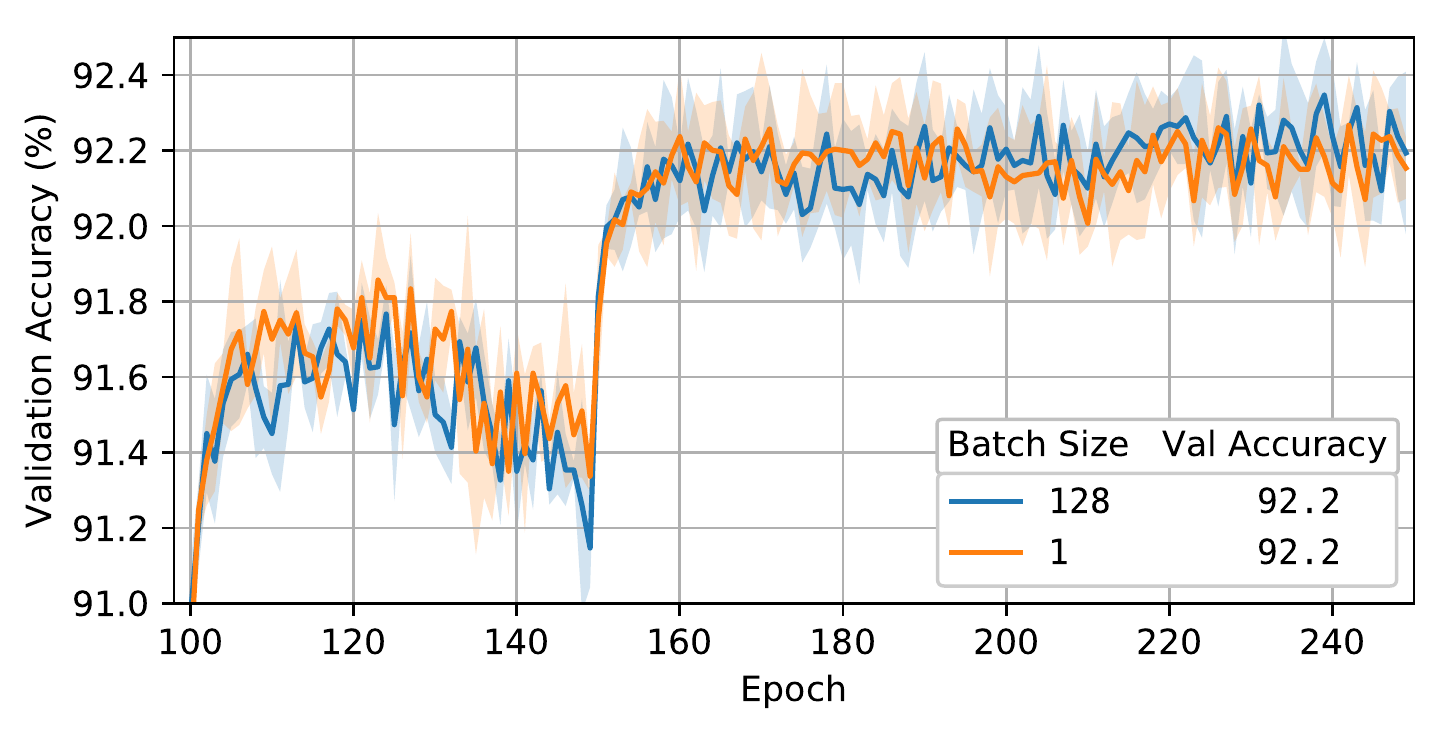}
        \caption{SGDM training of CIFAR-10 ResNet-20 with ON validation curves (mean$\pm$std.dev of 3 runs). Hyperparameter scaling rules described in Appendix~\ref{sec:smallbt} produce equivalent training.}
        \label{fig:c10r20bsizeacc}
    \end{minipage}
    \hfill
    \begin{minipage}[b]{.48\textwidth}
        \centering
        \includegraphics[width=\linewidth]{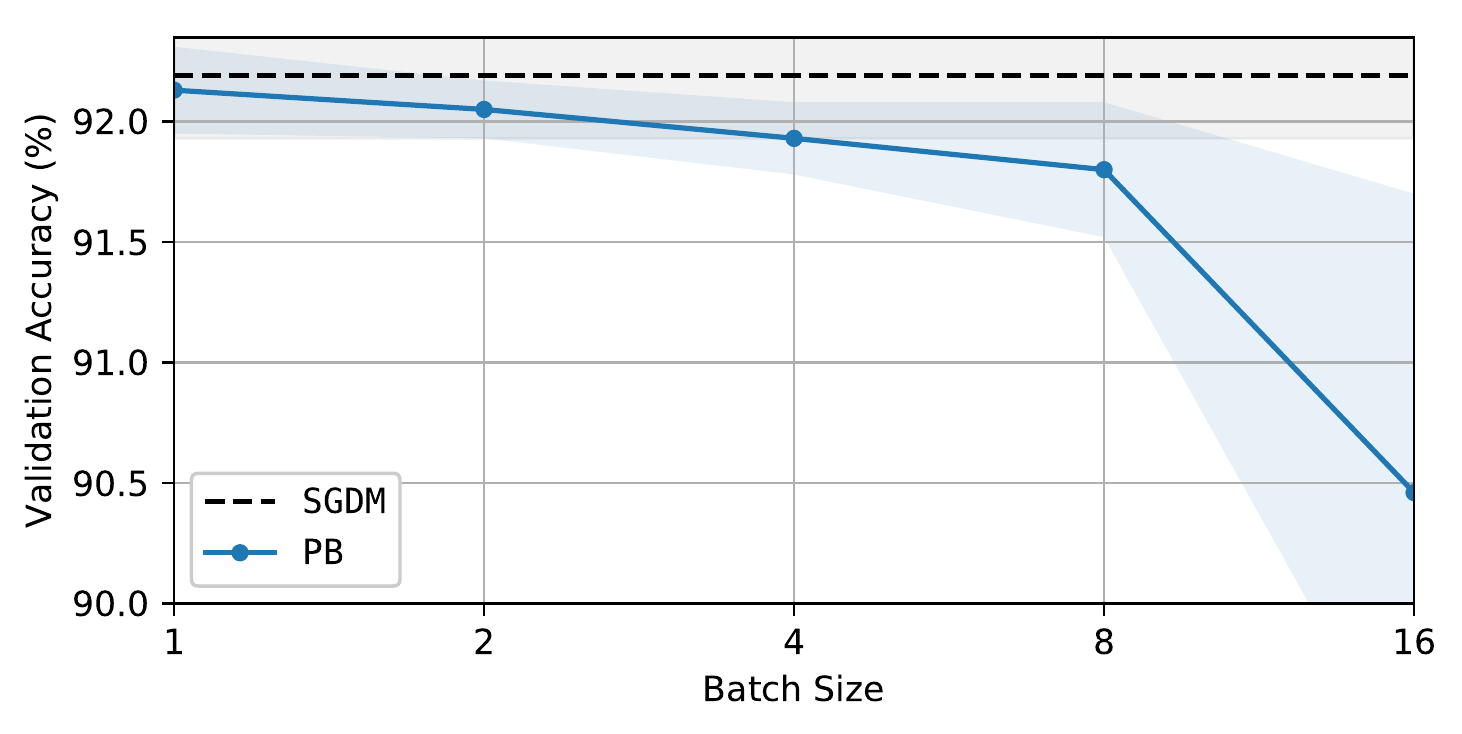}
        \caption{Pipelined Backpropagation training of CIFAR-10 ResNet-20 using Online Normalization. Final validation accuracy decreases as batch size is increased.}
        \label{fig:c10r20PBbsizeacc}
    \end{minipage}
\end{figure*}

\section{Experiments} 
\label{sec:experiments}

\begin{table}
    \caption{
        CIFAR-10 final validation accuracy (mean$\pm$std.dev of 3 runs) for ResNet (RN) using Online Normalization.
    }
    \label{tab:c10ONaccMeanRoundrtr}
    \centering
    \vskip 0.15in
    \small
    \sc
    \begin{tabular}{l|c|cc}
        \toprule
        Network & Stages & SGDM                    &  PB                     \\ \midrule
        RN20    & 34     & \textbf{92.19}$\pm$0.26 & \textbf{92.13}$\pm$0.18 \\
        RN32    & 52     & 92.86$\pm$0.04          & \textbf{93.05}$\pm$0.13 \\
        RN44    & 70     & 92.98$\pm$0.22          & \textbf{93.16}$\pm$0.13 \\
        RN56    & 88     & 93.21$\pm$0.07          & \textbf{93.37}$\pm$0.13 \\
        RN110   & 169    & \textbf{93.83}$\pm$0.11 & \textbf{93.74}$\pm$0.13 \\ \bottomrule 
    \end{tabular}
    \vskip -0.1in
\end{table}

The goal of our experiments is to investigate the convergence properties of small batch, fine-grained Pipeline Backpropagation and compare it to standard mini-batch gradient descent. 
As mentioned before, the experiments are performed on GPUs which do not benefit from pipelined training.
Therefore, we do not compare wall-time to convergence, but  \citeauthor{zhang2019scalable}~\citeyearpar{zhang2019scalable}, \citeauthor{li2017caterpillar}~\citeyearpar{li2017caterpillar}, and \citeauthor{chen2016eyeriss}~\citeyearpar{chen2016eyeriss}
have shown significant improvement in throughput and processing efficiency when pipelining neural network training on appropriate hardware.
We experiment with two families of networks, VGG~\cite{simonyan2014very} and pre-activation ResNets~\cite{he2016identity} on two commonly used image classification benchmarks, CIFAR-10~\cite{c10bib} and ImageNet~\cite{imagenet_cvpr09}. 
We adopt the data prepossessing and hyperparameter settings for VGG and ResNet from \citeauthor{fu_2019}~\citeyearpar{fu_2019} and \citeauthor{Chiley2019OnlineNF}~\citeyearpar{Chiley2019OnlineNF} respectively.
The delay created by PB is defined by the depth of the network architecture so we test various network depths. When applicable, each convolutional layer is combined with its associated activation function and normalization layer into a single pipeline stage. In our implementation the summation of a residual branch and a skip connection in residual networks also becomes a stage. 
In each row of Tables~\ref{tab:c10ONaccMeanRoundrtr}-\ref{tab:c10accSTWP}, the values within one standard error of the maximum accuracy are highlighted. 
Other details about our experimental setup can be found in Appendix~\ref{apdx:expdetails}.

\begin{figure}
    \centering
    \includegraphics[width=\linewidth]{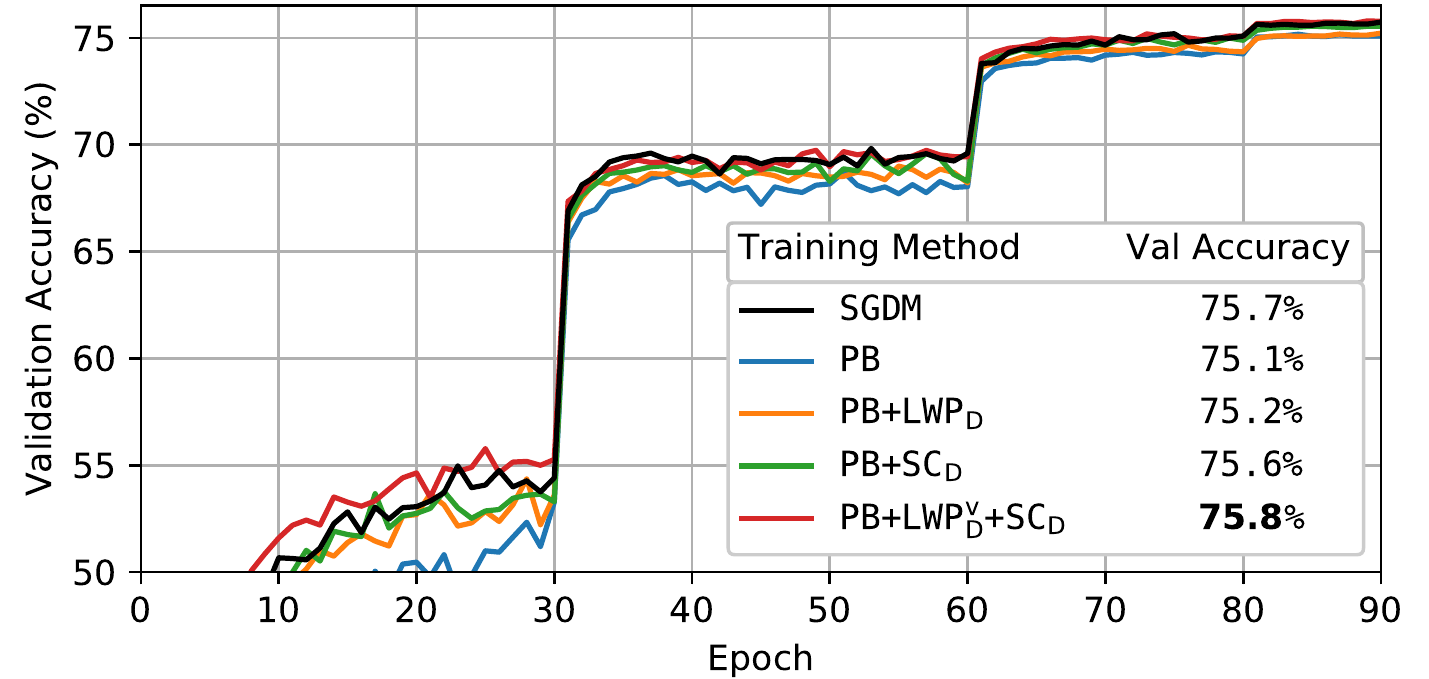}
    \caption{ImageNet ResNet-50 validation accuracy.}
    \label{fig:i1kr50acc}
\end{figure}

\subsection{Small Batch Sizes and Effective Normalization}
\label{sec:small_batch_sizes_and_online_normalization}
The memory requirements of pipeline parallelism have a quadratic dependency on the number of stages. To keep the overall memory requirements reasonable, we use a batch size of one in our PB experiments unless otherwise stated. We take the learning rate and momentum used at a reference batch size and scale them according to the rules provided in Appendix~\ref{sec:smallbt}. The rules attempt to keep the impulse response of each gradient (the contribution to weight updates over time) similar at different batch sizes. Figure~\ref{fig:c10r20bsizeacc} shows training curves for ResNet-20 on CIFAR-10 at a batch size of 1 and 128 when using Stochastic Gradient Descent with Momentum (SGDM). The curves are near identical, suggesting that the hyperparameter scaling rules produce similar training trajectories when using different batch sizes. This enables a fair comparison between PB and SGDM even though different batch sizes are used. For our SGDM baselines, we use a batch size of 128 for CIFAR-10 and 32 for ImageNet since SGDM training at batch size 1 on GPUs is slow and expensive. 

\begin{table*}
    \caption{
        CIFAR-10 final validation accuracy (mean$\pm$std.dev of 5 runs) for ResNet (RN) with group normalization and VGG training.
    }
    \label{tab:c10accMeanRoundrtr}
    \centering
    \vskip 0.15in
    \small
    \sc
    \begin{tabular}{l|c|cccccc}
        \toprule
        Network & Stages & SGDM                    &  PB                     & PipeDream                    & PB+LWP$_\mathrm{D}$     & PB+SC$_\mathrm{D}$      & PB+LWP$^\mathrm{v}_\mathrm{D}$+SC\textsubscript{D} \\ \midrule
        VGG11   & 29     & \textbf{91.16}$\pm$0.19 & 90.83$\pm$0.20          & 90.93$\pm$0.12               & 91.05$\pm$0.11          & \textbf{91.08}$\pm$0.19 & \textbf{91.12}$\pm$0.18                            \\
        VGG13   & 33     & \textbf{92.57}$\pm$0.15 & \textbf{92.59}$\pm$0.15 & 92.30$\pm$0.24               & 92.51$\pm$0.11          & 92.38$\pm$0.27          & \textbf{92.56}$\pm$0.14                            \\
        VGG16   & 39     & 92.24$\pm$0.19          & 92.06$\pm$0.21          & 59.31$\pm$45.01\footnotemark & 92.22$\pm$0.24          & \textbf{92.45}$\pm$0.30 & \textbf{92.38}$\pm$0.27                            \\ \midrule
        RN20    & 34     & 90.63$\pm$0.31          & 90.44$\pm$0.24          & 90.36$\pm$0.06               & 90.68$\pm$0.30          & \textbf{90.80}$\pm$0.29 & \textbf{90.92}$\pm$0.25                            \\
        RN32    & 52     & 91.68$\pm$0.23          & 91.46$\pm$0.09          & 91.40$\pm$0.28               & 91.66$\pm$0.10          & 91.55$\pm$0.14          & \textbf{92.04}$\pm$0.13                            \\
        RN44    & 70     & \textbf{92.19}$\pm$0.14 & 91.71$\pm$0.25          & 91.72$\pm$0.14               & 92.00$\pm$0.14          & \textbf{92.13}$\pm$0.16 & \textbf{92.16}$\pm$0.26                            \\
        RN56    & 88     & 92.39$\pm$0.20          & 91.89$\pm$0.40          & 91.82$\pm$0.19               & 92.31$\pm$0.14          & 92.33$\pm$0.16          & \textbf{92.48}$\pm$0.11                            \\
        RN110   & 169    & \textbf{92.77}$\pm$0.22 & 91.81$\pm$0.15          & 91.92$\pm$0.33               & \textbf{92.76}$\pm$0.05 & 92.28$\pm$0.29          & 92.41$\pm$0.16                                     \\ \bottomrule 
    \end{tabular}
    \vskip -0.1in
\end{table*}

In PB the delay (number of optimization steps) between the forward and backwards passes for a given stage is determined by the network architecture. The effect of the delay depends on the total weight change that occurs over the course of the delay. 
When the learning rate and momentum are scaled as we do, the magnitude of the weight change depends on the number of samples processed between the forward and backwards passes.
As a result, decreasing the batch size produce smaller changes in the weights and therefore helps mitigate the effects of delay and weight inconsistency in PB. Figure~\ref{fig:c10r20PBbsizeacc} shows that PB performs best at batch size one.

\begin{table}
    \caption{
        CIFAR-10 validation accuracy (mean$\pm$std.dev of 5 runs) when tuning the learning rate (LR) for ResNet-20 with GN training. The learning rate shown is used for batch size 128 SGDM training and is adjusted for batch size one PB training.
    }
    \label{tab:c10r20acclr}
    \centering
    \vskip 0.15in
    \small
    \sc
    \begin{tabular}{l|ccc}
        \toprule
        LR        & SGDM            &  PB            & $\!\!\!\!$PB+LWP$^\mathrm{v}_\mathrm{D}$+SC\textsubscript{D} \\ \midrule
        0.0125$\!\!$& 88.76$\pm$0.45  & 88.77$\pm$0.22 & \textbf{89.32}$\pm$0.26 \\
        0.025     & 89.88$\pm$0.32  & 89.55$\pm$0.35 & \textbf{90.06}$\pm$0.23 \\
        0.05      & 90.47$\pm$0.22  & 90.10$\pm$0.40 & \textbf{90.80}$\pm$0.37 \\
        0.1       & 90.63$\pm$0.31  & 90.44$\pm$0.24 & \textbf{90.92}$\pm$0.25 \\
        0.2       & 90.69$\pm$0.25  & 90.22$\pm$0.11 & \textbf{90.89}$\pm$0.28 \\
        0.4       & 89.54$\pm$0.32  & 88.82$\pm$0.32 & \textbf{89.93}$\pm$0.20 \\
        0.8       & 69.16$\pm$33.08
        \footnotemark[2]$\!\!\!\!$& 83.53$\pm$1.39 & \textbf{88.01}$\pm$0.56 \\
        \bottomrule
    \end{tabular}
    \vskip -0.1in
\end{table}
\footnotetext{Unstable training.}

\begin{table*}
    \caption{
        CIFAR-10 (C10) validation accuracy (mean$\pm$std.dev of five runs) and ImageNet (I1k) validation accuracy (single run) comparing SpecTrain and our methods for ResNet (RN) and VGG training.
    }
    \label{tab:c10accSTWP}
    \centering
    \vskip 0.15in
    \small
    \sc
    \begin{tabular}{l|cc|cc}
        \toprule
        Networks(Dataset) & SGDM                    &  PB                     &  PB+LWP$^\mathrm{v}_\mathrm{D}$+SC\textsubscript{D} & SpecTrain     \\ \midrule
        VGG13 (C10)       & \textbf{92.57}$\pm$0.15 & \textbf{92.59}$\pm$0.15 & \textbf{92.56}$\pm$0.14                             & \textbf{92.49}$\pm$0.12 \\
        RN20 (C10)        & 90.63$\pm$0.31          & 90.44$\pm$0.24          & \textbf{90.92}$\pm$0.25                             & \textbf{90.93}$\pm$0.09 \\
        RN56 (C10)        & 92.39$\pm$0.20          & 91.89$\pm$0.40          & 92.48$\pm$0.11                                      & \textbf{92.72}$\pm$0.10 \\ \midrule
        RN50 (I1k)        & 75.7                    & 75.1                    & \textbf{75.8}                                       & 75.3 \\ 
        \bottomrule
    \end{tabular}
    \vskip -0.1in
\end{table*}

Batch normalization has been shown to perform poorly at small batch sizes~\cite{Singh2019}. We experiment with two alternatives, Group Normalization (GN) from \citeauthor{Wu_2018_ECCV}~\citeyearpar{Wu_2018_ECCV} and Online Normalization (ON) proposed by \citeauthor{Chiley2019OnlineNF}~\citeyearpar{Chiley2019OnlineNF}, which both work at batch size one. With ON we find that PB training does not cause a loss of accuracy when training ResNets on CIFAR-10 with a small batch size (Table~\ref{tab:c10ONaccMeanRoundrtr}). Group Normalization does not perform as well for CIFAR-10 in general, the baseline SGDM accuracy is significantly lower. Online Normalization and Batch Normalization may benefit from a regularization effect caused by cross-sample noise in the normalization statistics. The original Group Normalization work shows good performance on ImageNet where such a regularization effect might be less important. Aside from the baseline accuracy degradation, performing PB with GN also suffers from an additional degradation depending on the pipeline depth (Figure~\ref{fig:i1kr50acc}\footnote{\citeauthor{Wu_2018_ECCV}~\citeyearpar{Wu_2018_ECCV} report an accuracy of 75.9\%. They do this by extending and modifying the learning rate schedule we used which we adopted from \cite{He_2016_CVPR}.}, Table~\ref{tab:c10accMeanRoundrtr}). The choice of normalization can therefore have a significant impact on the performance of PB.

\begin{figure}
    \centering
    \includegraphics[width=\linewidth]{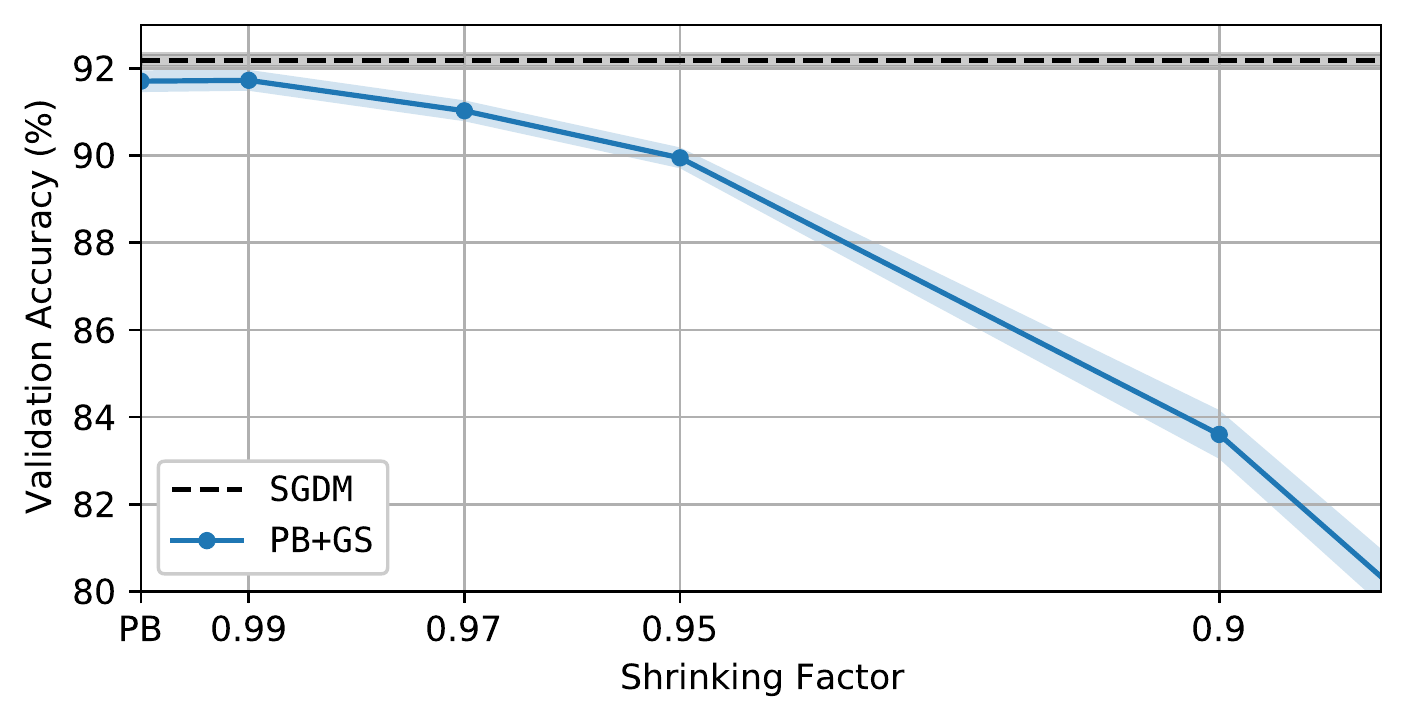}
    \caption{CIFAR-10 PB with Gradient Shrinking validation accuracy (mean$\pm$std.dev of 5 runs) for ResNet-44 with GN.}
    \label{fig:c10r20accGS}
\end{figure}

\subsection{Mitigation Methods}
Pipelined Backpropagation can suffer from a loss of accuracy or instability in some settings. This is more prominent in ResNets with GN and VGG networks so we use them to measure the effect of our mitigation methods and compare them to other existing methods found in literature. For LWP and SC the default hyperparameters suggested by our convex quadratic analysis (Section~\ref{sec:cq_analysis}) are used without further tuning. The results can potentially be improved with a hyperparameter search.

Table~\ref{tab:c10r20acclr} shows a learning rate sweep for training ResNet-20 with GN on CIFAR-10 with both SGDM and PB. We find that PB suffers from a small accuracy degradation that cannot be tuned away by adjusting the learning rate. The original learning rate of 0.1 adopted from \citeauthor{He_2016_CVPR}~\citeyearpar{He_2016_CVPR} and the appropriately scaled version for batch size one training is optimal in both cases.
We find that PB tolerates higher learning rates than SGDM. This could be an advantage of using small batches which \citeauthor{li2018visualizing}~\citeyearpar{li2018visualizing} suggest makes hyperparameters less sensitive than for larger batch sizes. Table~\ref{tab:c10r20acclr} shows our methods can improve convergence and
decrease sensitivity to changes in the learning rate.

Table~\ref{tab:c10accMeanRoundrtr} and Figure~\ref{fig:i1kr50acc} compare the accuracy of SGDM and PB training for several networks and different mitigation methods.
PB suffers from an accuracy degradation depending on the network depth.
Both Spike Compensation and Linear Weight Prediction help mitigate the loss of accuracy but their combination generally performs the best as suggested by our analysis (Section~\ref{sec:cq_analysis}).
The combination recovers the full SGDM accuracy for ResNet-50 training and for every CIFAR-10 experiment except for the deepest network, ResNet-110.
In that case it still closes most of the gap but LWP$_\mathrm{D}$ performs better and produces results competitive with SGDM.
Overcompensating for the delays is thus helpful in most cases but may be less effective for the large delays encountered in the deepest networks.
Training with large delays may be more sensitive to individual predictions and corrections, benefiting the comparatively more conservative LWP$_\mathrm{D}$.

Like \citeauthor{Chen2018EfficientAR}~\citeyearpar{Chen2018EfficientAR}, we find that the Weight Stashing method used in PipeDream~\cite{Harlap2018PipeDreamFA} to address weight inconsistency does not aid our training (Table~\ref{tab:c10accMeanRoundrtr}, Appendix~\ref{sec:pipemare}). In Appendix~\ref{sec:InconsistencyVsDelay} we show that the effects of weight inconsistency only become significant at large delays that are not encountered in our setting. Figure~\ref{fig:c10r20accGS} shows that Gradient Shrinking~\cite{Zhuang2019FullyDN} is also not effective for this type of training, no value of the shrinking factor is able to improve PB accuracy for RN44. Out of existing methods, SpecTrain~\cite{Chen2018EfficientAR} performs the best (Table~\ref{tab:c10accSTWP}). Similar to LWP$^\mathrm{v}_\mathrm{D}$+SC\textsubscript{D} it is able to recover or even improve accuracy on CIFAR-10. However, it is not able to recover accuracy for ResNet-50 training on ImageNet and doubles the memory and compute overhead of our combined mitigation (Appendix~\ref{sec:methods_overhead}). Out of the mitigation methods tested, LWP$^\mathrm{v}_\mathrm{D}$+SC\textsubscript{D} is thus the only one that is competitive to SGDM training in both the CIFAR-10 and ImageNet experiments.


\section{Conclusion}
\label{sec:discussion}
Fine-grained Pipelined Backpropagation has several advantages for hardware compared to traditional training using batch parallel Stochastic Gradient Descent. As discussed in Sections~\ref{sec:intro}~and~\ref{sec:fgpponHW} this can give large efficiency improvements for hardware architectures that can properly exploit these properties, such as Coarse-Grained Reconfigurable Arrays. However, traditional PB training can suffer from accuracy degradation and instability compared to standard training due to delayed gradients and weight inconsistency.

We show that small micro-batch sizes are crucial for making fine-grained Pipelined Backpropagation viable. Combined with an appropriate choice (or scaling) of hyperparameters, small batches reduce the negative effects of gradient delay and weight inconsistency. The use of small micro-batches also reduces the memory requirements that could otherwise be excessive. Unlike traditional training, fine-grained pipelined backpropagation can be efficient with small micro-batch sizes when combined with persistent kernels that do not need to amortize weight loading.

A good choice of normalization can also significantly aid Pipelined Backpropagation training. We experiment with two normalization options that work at batch size one, Online Normalization and Group Normalization. We observe that ON is significantly more robust to and helps stabilize training.

For cases where fine-grained, small batch, Pipelined Backpropagation suffers from an accuracy degradation, we present two new mitigation approaches, SC and LWP. We analyze their workings on a quadratic model which suggests that the methods can increase stability, accelerate convergence, and restore the beneficial effects of momentum in the presence of gradient delays. The analysis also suggests that combining the methods and thus “overcompensating” for the delays can improve convergence. Our neural network experiments with PB confirm these advantages. We find that the combined mitigation outperforms existing mitigation strategies, allowing our PB training to match the reference accuracy on both ImageNet and CIFAR-10 with minimal overhead and without the need of additional hyperparameter tuning. 

With our methods, PB is a promising alternative to traditional training. Future hardware architectures could reap significant efficiency gains from using small batch size, fine-grained Pipelined Backpropagation.

\section*{Acknowledgements}
\label{sec:acknowledgments}
We are grateful to Vithursan Thangarasa, Ron Estrin, Natalia Vassilieva, and Dennis DeCoste for their feedback on the manuscript. 
We thank Min Xu for his help with the dataloader used in our GProp experiments and Chuan-Yung Tsai for insightful discussions.


\bibliography{pbatscale}

\begin{thebibliography}{61}
\providecommand{\natexlab}[1]{#1}
\providecommand{\url}[1]{\texttt{#1}}
\expandafter\ifx\csname urlstyle\endcsname\relax
  \providecommand{\doi}[1]{doi: #1}\else
  \providecommand{\doi}{doi: \begingroup \urlstyle{rm}\Url}\fi

\bibitem[Amodei \& Hernandez(2018)Amodei and Hernandez]{amodei2018ai}
Amodei, D. and Hernandez, D.
\newblock {AI} and compute, 2018.
\newblock URL \url{https://openai.com/blog/ai-and-compute/}.

\bibitem[Avron et~al.(2015)Avron, Druinsky, and Gupta]{avron2015revisiting}
Avron, H., Druinsky, A., and Gupta, A.
\newblock Revisiting asynchronous linear solvers: Provable convergence rate
  through randomization.
\newblock \emph{Journal of the ACM (JACM)}, 62\penalty0 (6):\penalty0 51, 2015.

\bibitem[Belilovsky et~al.(2019)Belilovsky, Eickenberg, and
  Oyallon]{belilovsky2019decoupled}
Belilovsky, E., Eickenberg, M., and Oyallon, E.
\newblock Decoupled greedy learning of cnns.
\newblock \emph{arXiv preprint arXiv:1901.08164}, 2019.

\bibitem[Chen et~al.(2018)Chen, Yang, and Cheng]{Chen2018EfficientAR}
Chen, C.-C., Yang, C.-L., and Cheng, H.-Y.
\newblock Efficient and robust parallel dnn training through model parallelism
  on multi-gpu platform.
\newblock \emph{ArXiv}, abs/1809.02839, 2018.

\bibitem[Chen et~al.(2012)Chen, Eversole, Li, Yu, and Seide]{chen2012pipelined}
Chen, X., Eversole, A., Li, G., Yu, D., and Seide, F.
\newblock Pipelined back-propagation for context-dependent deep neural
  networks.
\newblock In \emph{Interspeech}. ISCA, September 2012.

\bibitem[Chen et~al.(2016)Chen, Emer, and Sze]{chen2016eyeriss}
Chen, Y.-H., Emer, J., and Sze, V.
\newblock Eyeriss: A spatial architecture for energy-efficient dataflow for
  convolutional neural networks.
\newblock \emph{ACM SIGARCH Computer Architecture News}, 44\penalty0
  (3):\penalty0 367--379, 2016.

\bibitem[Chen et~al.(2019)Chen, Yang, Emer, and Sze]{chen2019eyeriss}
Chen, Y.-H., Yang, T.-J., Emer, J., and Sze, V.
\newblock Eyeriss v2: A flexible accelerator for emerging deep neural networks
  on mobile devices.
\newblock \emph{IEEE Journal on Emerging and Selected Topics in Circuits and
  Systems}, 9\penalty0 (2):\penalty0 292--308, 2019.

\bibitem[Chetlur et~al.(2014)Chetlur, Woolley, Vandermersch, Cohen, Tran,
  Catanzaro, and Shelhamer]{chetlur2014cudnn}
Chetlur, S., Woolley, C., Vandermersch, P., Cohen, J., Tran, J., Catanzaro, B.,
  and Shelhamer, E.
\newblock {cuDNN}: Efficient primitives for deep learning.
\newblock \emph{arXiv preprint arXiv:1410.0759}, 2014.

\bibitem[Chiley et~al.(2019)Chiley, Sharapov, Kosson, Koster, Reece, Samaniego
  de~la Fuente, Subbiah, and James]{Chiley2019OnlineNF}
Chiley, V., Sharapov, I., Kosson, A., Koster, U., Reece, R., Samaniego de~la
  Fuente, S., Subbiah, V., and James, M.
\newblock Online normalization for training neural networks.
\newblock In \emph{Advances in Neural Information Processing Systems 32}, pp.\
  8431--8441. Curran Associates, Inc., 2019.

\bibitem[Chiley et~al.(2020)Chiley, Kosson, and Koster]{chiley2020error}
Chiley, V., Kosson, A., and Koster, U.
\newblock Error compensation mechanism in online normalization, Apr 2020.
\newblock URL
  \url{https://www.cerebras.net/error-compensation-mechanism-in-online-normalization/}.

\bibitem[Dauphin \& Schoenholz(2019)Dauphin and
  Schoenholz]{dauphin2019metainit}
Dauphin, Y.~N. and Schoenholz, S.
\newblock {MetaInit: Initializing learning by learning to initialize}.
\newblock In \emph{Advances in Neural Information Processing Systems}, pp.\
  12624--12636, 2019.

\bibitem[De \& Smith(2020)De and Smith]{de2020batch}
De, S. and Smith, S.~L.
\newblock Batch normalization biases deep residual networks towards shallow
  paths.
\newblock \emph{arXiv preprint arXiv:2002.10444}, 2020.

\bibitem[Deng et~al.(2009)Deng, Dong, Socher, Li, Li, and
  Fei-Fei]{imagenet_cvpr09}
Deng, J., Dong, W., Socher, R., Li, L.-J., Li, K., and Fei-Fei, L.
\newblock {ImageNet: A Large-Scale Hierarchical Image Database}.
\newblock In \emph{CVPR09}, 2009.

\bibitem[Diamos et~al.(2016)Diamos, Sengupta, Catanzaro, Chrzanowski, Coates,
  Elsen, Engel, Hannun, and Satheesh]{diamos16persistent}
Diamos, G., Sengupta, S., Catanzaro, B., Chrzanowski, M., Coates, A., Elsen,
  E., Engel, J., Hannun, A., and Satheesh, S.
\newblock Persistent rnns: Stashing recurrent weights on-chip.
\newblock In Balcan, M.~F. and Weinberger, K.~Q. (eds.), \emph{Proceedings of
  The 33rd International Conference on Machine Learning}, volume~48 of
  \emph{Proceedings of Machine Learning Research}, pp.\  2024--2033, New York,
  New York, USA, 20--22 Jun 2016. PMLR.

\bibitem[Fu(2019)]{fu_2019}
Fu, C.-Y.
\newblock pytorch-vgg-cifar10, May 2019.
\newblock URL \url{https://github.com/chengyangfu/pytorch-vgg-cifar10}.

\bibitem[Gaunt et~al.(2017)Gaunt, Johnson, Riechert, Tarlow, Tomioka,
  Vytiniotis, and Webster]{gaunt2017ampnet}
Gaunt, A.~L., Johnson, M.~A., Riechert, M., Tarlow, D., Tomioka, R.,
  Vytiniotis, D., and Webster, S.
\newblock {AMPNet}: Asynchronous model-parallel training for dynamic neural
  networks.
\newblock \emph{arXiv preprint arXiv:1705.09786}, 2017.

\bibitem[Giladi et~al.(2019)Giladi, Nacson, Hoffer, and
  Soudry]{giladi2019stability}
Giladi, N., Nacson, M.~S., Hoffer, E., and Soudry, D.
\newblock At stability's edge: How to adjust hyperparameters to preserve minima
  selection in asynchronous training of neural networks?
\newblock \emph{arXiv preprint arXiv:1909.12340}, 2019.

\bibitem[Goh(2017)]{goh2017why}
Goh, G.
\newblock Why momentum really works.
\newblock \emph{Distill}, 2017.
\newblock \doi{10.23915/distill.00006}.

\bibitem[{Graphcore}(2020)]{graphcore2020doc}
{Graphcore}.
\newblock Graphcore documents, 2020.
\newblock URL
  \url{https://docs.graphcore.ai/projects/tf-model-parallelism/en/latest/pipelining.html}.

\bibitem[Hakimi et~al.(2019)Hakimi, Barkai, Gabel, and
  Schuster]{hakimi2019taming}
Hakimi, I., Barkai, S., Gabel, M., and Schuster, A.
\newblock Taming momentum in a distributed asynchronous environment.
\newblock \emph{arXiv preprint arXiv:1907.11612}, 2019.

\bibitem[Harlap et~al.(2018)Harlap, Narayanan, Phanishayee, Seshadri, Devanur,
  Ganger, and Gibbons]{Harlap2018PipeDreamFA}
Harlap, A., Narayanan, D., Phanishayee, A., Seshadri, V., Devanur, N.~R.,
  Ganger, G.~R., and Gibbons, P.~B.
\newblock Pipe{D}ream: Fast and efficient pipeline parallel dnn training.
\newblock \emph{ArXiv}, abs/1806.03377, 2018.

\bibitem[He et~al.(2016{\natexlab{a}})He, Zhang, Ren, and Sun]{He_2016_CVPR}
He, K., Zhang, X., Ren, S., and Sun, J.
\newblock Deep residual learning for image recognition.
\newblock In \emph{The IEEE Conference on Computer Vision and Pattern
  Recognition (CVPR)}, June 2016{\natexlab{a}}.

\bibitem[He et~al.(2016{\natexlab{b}})He, Zhang, Ren, and Sun]{he2016identity}
He, K., Zhang, X., Ren, S., and Sun, J.
\newblock Identity mappings in deep residual networks.
\newblock In \emph{European conference on computer vision}, pp.\  630--645.
  Springer, 2016{\natexlab{b}}.

\bibitem[Huang et~al.(2018)Huang, Cheng, Chen, Lee, Ngiam, Le, and
  Chen]{Huang2018GPipeET}
Huang, Y., Cheng, Y., Chen, D., Lee, H., Ngiam, J., Le, Q.~V., and Chen, Z.
\newblock {GPipe}: Efficient training of giant neural networks using pipeline
  parallelism.
\newblock \emph{ArXiv}, abs/1811.06965, 2018.

\bibitem[Huo et~al.(2018{\natexlab{a}})Huo, Gu, and Huang]{huo2018training}
Huo, Z., Gu, B., and Huang, H.
\newblock Training neural networks using features replay.
\newblock In Bengio, S., Wallach, H., Larochelle, H., Grauman, K.,
  Cesa-Bianchi, N., and Garnett, R. (eds.), \emph{Advances in Neural
  Information Processing Systems 31}, pp.\  6659--6668. Curran Associates,
  Inc., 2018{\natexlab{a}}.

\bibitem[Huo et~al.(2018{\natexlab{b}})Huo, Gu, qian Yang, and
  Huang]{huo2018decoupled}
Huo, Z., Gu, B., qian Yang, and Huang, H.
\newblock Decoupled parallel backpropagation with convergence guarantee.
\newblock In Dy, J. and Krause, A. (eds.), \emph{Proceedings of the 35th
  International Conference on Machine Learning}, volume~80 of \emph{Proceedings
  of Machine Learning Research}, pp.\  2098--2106, Stockholmsmässan, Stockholm
  Sweden, 10--15 Jul 2018{\natexlab{b}}. PMLR.

\bibitem[Ioffe \& Szegedy(2015)Ioffe and
  Szegedy]{Ioffe:2015:BNA:3045118.3045167}
Ioffe, S. and Szegedy, C.
\newblock Batch normalization: Accelerating deep network training by reducing
  internal covariate shift.
\newblock In \emph{Proceedings of the 32nd International Conference on
  International Conference on Machine Learning - Volume 37}, ICML’15, pp.\
  448–456. JMLR.org, 2015.

\bibitem[Jaderberg et~al.(2017)Jaderberg, Czarnecki, Osindero, Vinyals, Graves,
  Silver, and Kavukcuoglu]{jaderberg2017decoupled}
Jaderberg, M., Czarnecki, W.~M., Osindero, S., Vinyals, O., Graves, A., Silver,
  D., and Kavukcuoglu, K.
\newblock Decoupled neural interfaces using synthetic gradients.
\newblock In Precup, D. and Teh, Y.~W. (eds.), \emph{Proceedings of the 34th
  International Conference on Machine Learning}, volume~70 of \emph{Proceedings
  of Machine Learning Research}, pp.\  1627--1635, International Convention
  Centre, Sydney, Australia, 06--11 Aug 2017. PMLR.

\bibitem[Jia et~al.(2019)Jia, Tillman, Maggioni, and
  Scarpazza]{jia2019dissecting}
Jia, Z., Tillman, B., Maggioni, M., and Scarpazza, D.~P.
\newblock Dissecting the graphcore ipu architecture via microbenchmarking.
\newblock \emph{arXiv preprint arXiv:1912.03413}, 2019.

\bibitem[Krizhevsky et~al.(2009)Krizhevsky, Hinton, et~al.]{c10bib}
Krizhevsky, A., Hinton, G., et~al.
\newblock Learning multiple layers of features from tiny images, 2009.

\bibitem[Li et~al.(2018)Li, Xu, Taylor, Studer, and
  Goldstein]{li2018visualizing}
Li, H., Xu, Z., Taylor, G., Studer, C., and Goldstein, T.
\newblock Visualizing the loss landscape of neural nets.
\newblock In \emph{Advances in Neural Information Processing Systems}, pp.\
  6389--6399, 2018.

\bibitem[Li \& Pedram(2017)Li and Pedram]{li2017caterpillar}
Li, Y. and Pedram, A.
\newblock Caterpillar: Coarse grain reconfigurable architecture for
  accelerating the training of deep neural networks.
\newblock In \emph{2017 IEEE 28th International Conference on
  Application-specific Systems, Architectures and Processors (ASAP)}, pp.\
  1--10. IEEE, 2017.

\bibitem[Lian et~al.(2015)Lian, Huang, Li, and Liu]{lian2015asynchronous}
Lian, X., Huang, Y., Li, Y., and Liu, J.
\newblock Asynchronous parallel stochastic gradient for nonconvex optimization.
\newblock In \emph{Advances in Neural Information Processing Systems}, pp.\
  2737--2745, 2015.

\bibitem[Lie(2020)]{lie2020wafer}
Lie, S.
\newblock Wafer-scale ml, 2 2020.
\newblock URL \url{https://info.matroid.com/scaledml-media-archive-2020}.

\bibitem[Masters \& Luschi(2018)Masters and Luschi]{Masters2018RevisitingSB}
Masters, D. and Luschi, C.
\newblock Revisiting small batch training for deep neural networks.
\newblock \emph{ArXiv}, abs/1804.07612, 2018.

\bibitem[Mattson et~al.(2020)Mattson, Cheng, Diamos, Coleman, Micikevicius,
  Patterson, Tang, Wei, Bailis, Bittorf, Brooks, Chen, Dutta, Gupta, Hazelwood,
  Hock, Huang, Kang, Kanter, Kumar, Liao, Narayanan, Oguntebi, Pekhimenko,
  Pentecost, Janapa~Reddi, Robie, St~John, Wu, Xu, Young, and
  Zaharia]{MLSYS2020_02522a2b}
Mattson, P., Cheng, C., Diamos, G., Coleman, C., Micikevicius, P., Patterson,
  D., Tang, H., Wei, G.-Y., Bailis, P., Bittorf, V., Brooks, D., Chen, D.,
  Dutta, D., Gupta, U., Hazelwood, K., Hock, A., Huang, X., Kang, D., Kanter,
  D., Kumar, N., Liao, J., Narayanan, D., Oguntebi, T., Pekhimenko, G.,
  Pentecost, L., Janapa~Reddi, V., Robie, T., St~John, T., Wu, C.-J., Xu, L.,
  Young, C., and Zaharia, M.
\newblock Mlperf training benchmark.
\newblock In Dhillon, I., Papailiopoulos, D., and Sze, V. (eds.),
  \emph{Proceedings of Machine Learning and Systems}, volume~2, pp.\  336--349,
  2020.

\bibitem[Mitliagkas et~al.(2016)Mitliagkas, Zhang, Hadjis, and
  R{\'e}]{Mitliagkas2016AsynchronyBM}
Mitliagkas, I., Zhang, C., Hadjis, S., and R{\'e}, C.
\newblock Asynchrony begets momentum, with an application to deep learning.
\newblock \emph{2016 54th Annual Allerton Conference on Communication, Control,
  and Computing (Allerton)}, pp.\  997--1004, 2016.

\bibitem[Nicol(2017)]{nicol2017coarse}
Nicol, C.
\newblock A coarse grain reconfigurable array (cgra) for statically scheduled
  data flow computing, 2017.

\bibitem[NVIDIA(2020)]{nvidiaa1002020}
NVIDIA.
\newblock Nvidia a100 tensor core gpu architecture, 2020.
\newblock URL
  \url{https://www.nvidia.com/content/dam/en-zz/Solutions/Data-Center/nvidia-ampere-architecture-whitepaper.pdf}.

\bibitem[NVIDIA(2021)]{nvidia_2021}
NVIDIA.
\newblock Nvidia data center deep learning product performance, Jan 2021.
\newblock URL
  \url{https://developer.nvidia.com/deep-learning-performance-training-inference}.

\bibitem[O’donoghue \& Candes(2015)O’donoghue and Candes]{o2015adaptive}
O’donoghue, B. and Candes, E.
\newblock Adaptive restart for accelerated gradient schemes.
\newblock \emph{Foundations of computational mathematics}, 15\penalty0
  (3):\penalty0 715--732, 2015.

\bibitem[Paszke et~al.(2019)Paszke, Gross, Massa, Lerer, Bradbury, Chanan,
  Killeen, Lin, Gimelshein, Antiga, et~al.]{paszke2017automatic}
Paszke, A., Gross, S., Massa, F., Lerer, A., Bradbury, J., Chanan, G., Killeen,
  T., Lin, Z., Gimelshein, N., Antiga, L., et~al.
\newblock Pytorch: An imperative style, high-performance deep learning library.
\newblock In \emph{Advances in Neural Information Processing Systems}, pp.\
  8024--8035, 2019.

\bibitem[P{\'e}trowski et~al.(1993)P{\'e}trowski, Dreyfus, and
  Girault]{Ptrowski1993PerformanceAO}
P{\'e}trowski, A., Dreyfus, G., and Girault, C.
\newblock Performance analysis of a pipelined backpropagation parallel
  algorithm.
\newblock \emph{IEEE transactions on neural networks}, 4 6:\penalty0 970--81,
  1993.

\bibitem[Podobas et~al.(2020)Podobas, Sano, and Matsuoka]{podobas2020survey}
Podobas, A., Sano, K., and Matsuoka, S.
\newblock A survey on coarse-grained reconfigurable architectures from a
  performance perspective.
\newblock \emph{arXiv preprint arXiv:2004.04509}, 2020.

\bibitem[Qiao et~al.(2019)Qiao, Wang, Liu, Shen, and
  Yuille]{DBLP:journals/corr/abs-1903-10520}
Qiao, S., Wang, H., Liu, C., Shen, W., and Yuille, A.~L.
\newblock Weight standardization.
\newblock \emph{CoRR}, abs/1903.10520, 2019.

\bibitem[{Sambanova}(2020)]{sambanova2020whitepaper}
{Sambanova}.
\newblock Accelerated computing with a reconfigurable dataflow architecture,
  2020.
\newblock URL
  \url{https://sambanova.ai/wp-content/uploads/2020/12/RDA-Whitepaper.pdf}.

\bibitem[{Serebryakov} et~al.(2019){Serebryakov}, {Milojicic}, {Vassilieva},
  {Fleischman}, and {Clark}]{serebryakov2019deep}
{Serebryakov}, S., {Milojicic}, D., {Vassilieva}, N., {Fleischman}, S., and
  {Clark}, R.~D.
\newblock Deep learning cookbook: Recipes for your ai infrastructure and
  applications.
\newblock In \emph{2019 IEEE International Conference on Rebooting Computing
  (ICRC)}, pp.\  1--9, 2019.
\newblock \doi{10.1109/ICRC.2019.8914704}.

\bibitem[Shallue et~al.(2019)Shallue, Lee, Antognini, Sohl-Dickstein, Frostig,
  and Dahl]{shallue2019measuring}
Shallue, C.~J., Lee, J., Antognini, J., Sohl-Dickstein, J., Frostig, R., and
  Dahl, G.~E.
\newblock Measuring the effects of data parallelism on neural network training.
\newblock \emph{Journal of Machine Learning Research}, 20\penalty0
  (112):\penalty0 1--49, 2019.

\bibitem[Shoeybi et~al.(2019)Shoeybi, Patwary, Puri, LeGresley, Casper, and
  Catanzaro]{shoeybi2019megatron}
Shoeybi, M., Patwary, M., Puri, R., LeGresley, P., Casper, J., and Catanzaro,
  B.
\newblock Megatron-lm: Training multi-billion parameter language models using
  gpu model parallelism.
\newblock \emph{arXiv preprint arXiv:1909.08053}, 2019.

\bibitem[Simonyan \& Zisserman(2014)Simonyan and Zisserman]{simonyan2014very}
Simonyan, K. and Zisserman, A.
\newblock Very deep convolutional networks for large-scale image recognition.
\newblock \emph{arXiv preprint arXiv:1409.1556}, 2014.

\bibitem[Singh \& Krishnan(2019)Singh and Krishnan]{Singh2019}
Singh, S. and Krishnan, S.
\newblock Filter response normalization layer: Eliminating batch dependence in
  the training of deep neural networks.
\newblock \emph{arXiv preprint arXiv:1911.09737}, 2019.

\bibitem[Vassilieva(2020)]{vassilieva2020neural}
Vassilieva, N.
\newblock Neural network parallelism at wafer scale, Apr 2020.
\newblock URL
  \url{https://www.cerebras.net/data-model-pipeline-parallel-training-neural-networks/}.

\bibitem[Venigalla et~al.(2020)Venigalla, Kosson, Chiley, and
  K{\"o}ster]{venigalla2020adaptive}
Venigalla, A., Kosson, A., Chiley, V., and K{\"o}ster, U.
\newblock {Adaptive Braking for Mitigating Gradient Delay}.
\newblock In \emph{International Conference on Machine Learning Workshop on
  Beyond First-Order Optimization Methods in Machine Learning}, 2020.

\bibitem[Wu \& He(2018)Wu and He]{Wu_2018_ECCV}
Wu, Y. and He, K.
\newblock Group normalization.
\newblock In \emph{The European Conference on Computer Vision (ECCV)},
  September 2018.

\bibitem[Xu et~al.(2019)Xu, Huo, and Huang]{xu2019diversely}
Xu, A., Huo, Z., and Huang, H.
\newblock Diversely stale parameters for efficient training of cnns.
\newblock \emph{arXiv preprint arXiv:1909.02625}, 2019.

\bibitem[Yang et~al.(2019)Yang, Zhang, Li, R{\'e}, Aberger, and
  De~Sa]{yang2019pipemare}
Yang, B., Zhang, J., Li, J., R{\'e}, C., Aberger, C.~R., and De~Sa, C.
\newblock Pipemare: Asynchronous pipeline parallel {DNN} training.
\newblock \emph{arXiv preprint arXiv:1910.05124}, 2019.

\bibitem[Zhang et~al.(2019{\natexlab{a}})Zhang, Li, Nado, Martens, Sachdeva,
  Dahl, Shallue, and Grosse]{zhang2019algorithmic}
Zhang, G., Li, L., Nado, Z., Martens, J., Sachdeva, S., Dahl, G., Shallue, C.,
  and Grosse, R.~B.
\newblock Which algorithmic choices matter at which batch sizes? insights from
  a noisy quadratic model.
\newblock In \emph{Advances in Neural Information Processing Systems}, pp.\
  8196--8207, 2019{\natexlab{a}}.

\bibitem[Zhang et~al.(2019{\natexlab{b}})Zhang, Dauphin, and Ma]{Zhang2019}
Zhang, H., Dauphin, Y.~N., and Ma, T.
\newblock {Fixup initialization: Residual learning without normalization}.
\newblock In \emph{7th International Conference on Learning Representations,
  ICLR 2019}, pp.\  1--16, 2019{\natexlab{b}}.

\bibitem[Zhang \& Mitliagkas(2017)Zhang and Mitliagkas]{zhang2017yellowfin}
Zhang, J. and Mitliagkas, I.
\newblock Yellowfin and the art of momentum tuning.
\newblock \emph{arXiv preprint arXiv:1706.03471}, 2017.

\bibitem[Zhang et~al.(2019{\natexlab{c}})Zhang, Rucker, Vilim, Prabhakar,
  Hwang, and Olukotun]{zhang2019scalable}
Zhang, Y., Rucker, A., Vilim, M., Prabhakar, R., Hwang, W., and Olukotun, K.
\newblock Scalable interconnects for reconfigurable spatial architectures.
\newblock In \emph{2019 ACM/IEEE 46th Annual International Symposium on
  Computer Architecture (ISCA)}, pp.\  615--628. IEEE, 2019{\natexlab{c}}.

\bibitem[Zhuang et~al.(2019)Zhuang, Wang, Liu, and Lin]{Zhuang2019FullyDN}
Zhuang, H., Wang, Y., Liu, Q., and Lin, Z.
\newblock Fully decoupled neural network learning using delayed gradients.
\newblock \emph{ArXiv}, abs/1906.09108, 2019.

\end{thebibliography}
\bibliographystyle{mlsys2021}

\clearpage
\appendix


\section{Data Parallel Compute Efficiency}
\label{sec:DPefficiency}
The limited compute efficiency of large scale data parallel training systems is a major motivation for exploring alternative training methods.
We use the theoretical utilization to quantify the compute efficiency, defined as:
\begin{equation}
    \text{utilization} = \frac{\text{FLOPS}_\text{U}}{\text{FLOPS}_\text{H}} = 
    \frac{\frac{\text{FLOP}_\text{N}}{\text{sample}} \frac{\text{samples}}{\text{sec}}}{\text{FLOPS}_\text{H}}
\end{equation}
where $\text{FLOPS}_\text{U}$ is the number of useful floating point operations performed each second during training, $\text{FLOPS}_\text{H}$ denotes the rated maximum FLOPS for the hardware, and $\frac{\text{FLOP}_\text{N}}{\text{sample}}$ represents the number of floating point operations required to train the network on a single sample.
Nvidia publishes the training speed (measured in samples per second) for highly optimized implementations of various common neural networks \cite{nvidia_2021}.
Paired with the FLOPs used per sample \cite{serebryakov2019deep} and the hardware's max rated FLOPS \cite{nvidiaa1002020} we can calculate the compute utilization efficiency of training common neural networks using data parallelism on widely used deep learning hardware (Table~\ref{tab:gpu_utilization}).
Using a higher number of accelerators can reduce the wall-clock time of training, but generally lowers the compute efficiency as can be seen from MLPerf~\cite{MLSYS2020_02522a2b}.
The worker specialization that fine grained pipeline parallelism enables is promising for efficient training in terms of power, and throughput \cite{zhang2019scalable,li2017caterpillar,chen2016eyeriss,chen2019eyeriss}.
This work explores ways to mitigate the downsides of pipeline-parallel training making it even more efficient.

\begin{table}
    \caption{GPU FLOPS utilization for training neural networks.}
    \label{tab:gpu_utilization}
    \centering
    \vskip 0.15in
    \small
    \sc
    \begin{tabular}{l|cc|cc|cc}
        \toprule
        Network             & System   & \#GPUs & Utilization \\ \midrule
        ResNet50            & A100 DGX & 1      & 16.4\%      \\
        ResNet50            & A100 DGX & 8      & 15.9\%      \\
        BERT$_\text{Large}$ & A100 DGX & 8      & 36.8\%      \\
        \bottomrule
    \end{tabular}
    \vskip -0.1in
\end{table}


\section{Pipelined Backpropagation}
\label{apdx:pb}
Pipeline parallelism is an interesting alternative or supplement to data parallelism (Appendix~\ref{apdx:DPvPP}).
To perform SGD training using pipeline parallelism, the same weights must be used on the forward and backwards passes. To satisfy this the pipeline needs to be empty before updating the weights. While the pipeline is filling or draining some workers sit idle which lowers utilization. The fill and drain overhead is illustrated in Figure~\ref{fig:fdvspb}.

Assume a pipeline has $S$ pipeline stages and each stage performs a single forward and a single backward transformation at each time step. Each sample is processed in $2S-1$ time steps. Performing a mini-batch SGD update with $N$ samples takes roughly $N+2S-2 \approx N+2S$ steps
\footnote{This is assuming the workers are unable to speed up processing when they only perform one of the transformations, otherwise it may be about $N+S$.}.
The work performed only corresponds to $N$ fully utilized steps so the overall utilization is upper bounded by
\begin{equation}
    \frac{N}{N+2S}
\end{equation}
Unless $N \gg S$ this represents a significant overhead.

Pipelined Backpropagation~\cite{Ptrowski1993PerformanceAO} avoids the fill and drain overhead by relaxing the constraint that the same weights must be used for the forward and backwards passes. In PB the pipeline is not drained before an update is applied, instead the parameters are updated as soon as $N$ gradients have been obtained. This keeps all workers utilized after the pipeline is filled for the first time (Figure \ref{fig:fdvspb}). We assume a batch size ($N$) of one and scale hyperparameters appropriately (Appendix~\ref{sec:smallbt}). We compare the weight updates of PB and SGD. We write SGD as:
\begin{equation}
    \theta_{t+1} = \theta_t - \eta \nabla L(x_t ; \theta_t)
    \label{eqn:sgd}
\end{equation}
where $\theta$ is the set of all model weights, $x_t$ is the sample at time $t$, $\eta$ is the learning rate, and $L$ is the loss function.
For PB we define $w^s_i$ to be the weights for pipeline stage $s\in [0,...,S-1]$ as seen by the $i^{\text{th}}$ sample, $x_i$, as it propagates backwards through the network.
$W_{i}$ is defined as the concatenation (denoted by $||$) of $w^s_i$ for all stages and corresponds to the weights on the blue line in Figure~\ref{fig:fdvspb}:
\begin{equation}
    W_i =  w_{i}^0\  ||\  w_{i}^1\  ||\  \cdots \ ||\  w_{i}^{S-1}
    \label{eqn:wi}
\end{equation}
The weight update for $x_i$ can then be written as:
\begin{equation}
    W_{i+1} = W_i - \eta G(x_i ; F_i, W_i)
    \label{eqn:sgdpb}
\end{equation}
where $G$ approximates the gradient and $F_i$ is the network state used for the forward pass of the network and corresponds to the weights on the red line in Figure~\ref{fig:fdvspb}.
For PB with $N=1$:
\begin{equation}
    F_i =  w_{i-2(S-1)}^0 \ ||\   w_{i-2(S-2)}^1\  ||\  \cdots\  ||\  w_{i}^{S-1}
    \label{eqn:wf}
\end{equation}
\eqnref{eqn:wi} - \eqnref{eqn:wf} reveal that PB differs from SGD in two ways: \textit{inconsistent weights} and \textit{stale gradients}.

\textbf{Inconsistent Weights}
\label{inconsistentweights}
Different weight are used during the forward and backwards pass, $W_i \ne F_i$. The resulting sample gradient is not the true sample gradient.
The inconsistency is greater for earlier stages in the pipeline.
If Weight Stashing~\cite{Harlap2018PipeDreamFA} is used to mitigate weight inconsistency the update is:
\begin{equation}
    W_{i+1} = W_i - \eta G(x_i ; F_i,  F_i) = W_i - \eta \nabla L(x_i ;  F_i)
    \label{eqn:sgdpb_weight_stash}
\end{equation}
Weight Stashing (WS) requires the overhead of storing parameter versions along with the activations.

\textbf{Stale Gradients}
In PB each gradient is obtained using weights from various time steps. 
When the gradient is obtained the weights have been updated.
This results in stale gradients (aka. \textit{delayed gradients}), an issue that also occurs in asynchronous SGD training \cite{lian2015asynchronous,avron2015revisiting}.
The gradient staleness varies by stage, earlier stages suffer from a greater degree of staleness.
The length of the grey lines in Figure~\ref{fig:fdvspb} is proportional to the age of the weights, which is also a measure of the gradient delay for each stage.
The depth of the pipeline determines the maximum delay. 
Weight Stashing does not address gradient delay because $F_i$ in \eqnref{eqn:sgdpb_weight_stash} is a delayed version of $W_i$.


\section{Batch Parallel vs Pipeline Parallel Computation}
\label{apdx:DPvPP}

Pipeline parallelism differs from batch parallelism in several ways:
\begin{itemize}[noitemsep,topsep=0pt]
    \item The training memory requirements differ. In both cases we assume an $L$ layer network trained with $W$ workers. During neural network training, the activations of many layers must be stored for the gradient calculation. For batch parallelism the activation memory required is $O(LW)$. To compute the backwards pass, each worker has to store activations for roughly every layer.
    In the pipeline parallel setting, each worker is responsible for storing the activations of approximately $L/W$ layers. The first worker must store its activations for $~2W$ steps. 
    The second worker needs to keep activations for $~2(W-1)$ steps and so on. 
    The total activation memory comes out to be approximately the same, $O(LW)$, however, the per worker memory requirements can be very different.
    Pipeline parallelism generally requires less memory for storing model parameters potentially requiring only a single copy of each parameter. Unless special methods are used, batch parallelism may need to keep $W$ copies of the model.
    \item The communication pattern is different. In pipeline parallelism each worker sends activations and the corresponding gradients to their neighbors. In distributed mini-batch training every worker must send the gradients for all model parameters and receive updated values after every batch. The bandwidth requirements in each case depend on the exact model used, the batch size, as well as other factors.
    \item Both pipeline parallel training and synchronized distributed batch parallel training can suffer from worker balancing bottlenecks. When using pipeline parallelism, care must be taken to balance the throughput of all workers since the overall speed is determined by the slowest worker. This load balancing issue could be handled in software \cite{Harlap2018PipeDreamFA} without requiring users to manually specify the model division. In synchronized distributed SGD care must be taken to balance the throughput and master node communication of all workers since the overall speed is determined by the slowest worker. 
    \item Batch normalization~\cite{Ioffe:2015:BNA:3045118.3045167} requires batch parallelism. In our work we are interested in replacing batch parallelism with fine-grained pipeline parallelism. We therefore operate at a micro-batch size of one which does not work well with Batch Normalization.
    Newer normalization techniques such as Group Normalization~\cite{Wu_2018_ECCV}, Weight Standardization~\cite{DBLP:journals/corr/abs-1903-10520}, Filter Response Normalization~\cite{Singh2019} and Online Normalization~\cite{Chiley2019OnlineNF,chiley2020error} are alternative normalization techniques which work well and can be used with small batch sizes. Alternatively initialization methods can be used to enable training without normalization \cite{Zhang2019,dauphin2019metainit,de2020batch}.
\end{itemize}

\section{GPU Bottlenecks for Fine-Grained Pipelined Training}
\label{sec:gpu_pp_lim}

A GPU consists of multiple cores with limited local memory and large off-chip main memory.
The main memory has high latency and relatively low bandwidth compared to the arithmetic throughput of the cores.
To achieve high compute utilization, a GPU kernel must be designed to work around these memory bandwidth and latency limitations.
Latency limitations are often managed by running a large number of threads in parallel and context switching to another thread when waiting on memory.
Large batch operations can provide a workload with enough threads to hide latency issues but the arithmetic intensity of operations must also be high enough to alleviate bandwidth bottlenecks.
Without sufficient cache reuse bandwidth becomes an issue. This can happen with small batch sizes and certain operations such as depth-wise convolutions, element-wise functions and sparse matrix multiplications.

Due to the activation memory requirements of pipelined training, modern network using this paradigm must use small batch sizes to fit in GPU memory.
At small batch sizes, the amount of computation per kernel might be insufficient to utilize all compute resources, but a large number of kernels can run in parallel which can significantly increase compute utilization.
The compute throughput of the GPU is equal to the rate at which kernels are launched multiplied by the work done by each kernel. 
As the work per kernel is decreased, the kernel launch rate must be increased to maintain compute throughput. 
At batch size one it may not be possible to launch kernels at a sufficient rate to keep utilization high on small networks such as ResNet-20 for CIFAR-10.
Larger networks tend to have more work done per kernel and do not suffer from this problem.

Without significant weight reuse, GPUs become memory bandwidth limited. 
For convolutional layers the weights are reused over the spatial and batch dimensions. 
Weight reuse increases as the spatial dimensions of the inputs increase. 
This makes bandwidth less of an issue for ImageNet (i1k) scale networks when compared to CIFAR-10 scale networks.

There are a few other challenges to small batch sized training. 
At small batch sizes optimizer overheads become significant. 
Each optimizer step requires loading the entire model, consuming significant memory bandwidth. 
At large batch sizes this is amortized over the batch size. 
For a batch size of one the optimizer steps consume a large fraction of the total memory bandwidth. 
Similarly, the time required for any new memory allocations cannot be amortized over the batch size. 
While persistent kernels can enable some of the advantages and speedups of worker specialization \cite{diamos16persistent}, the total on-chip memory of modern GPUs is too limited to run large models using fine-grained pipelined paralellism with persistent kernels; running multiple kernels on multiple threads concurrently limits the resources available to each threads making it impossible to keep weights in local memory or use persistent kernels for our work.


\begin{figure*}[ht]
     \centering
     \begin{subfigure}[b]{0.48\linewidth}
         \centering
         \includegraphics[width=\linewidth,height=4.7cm]{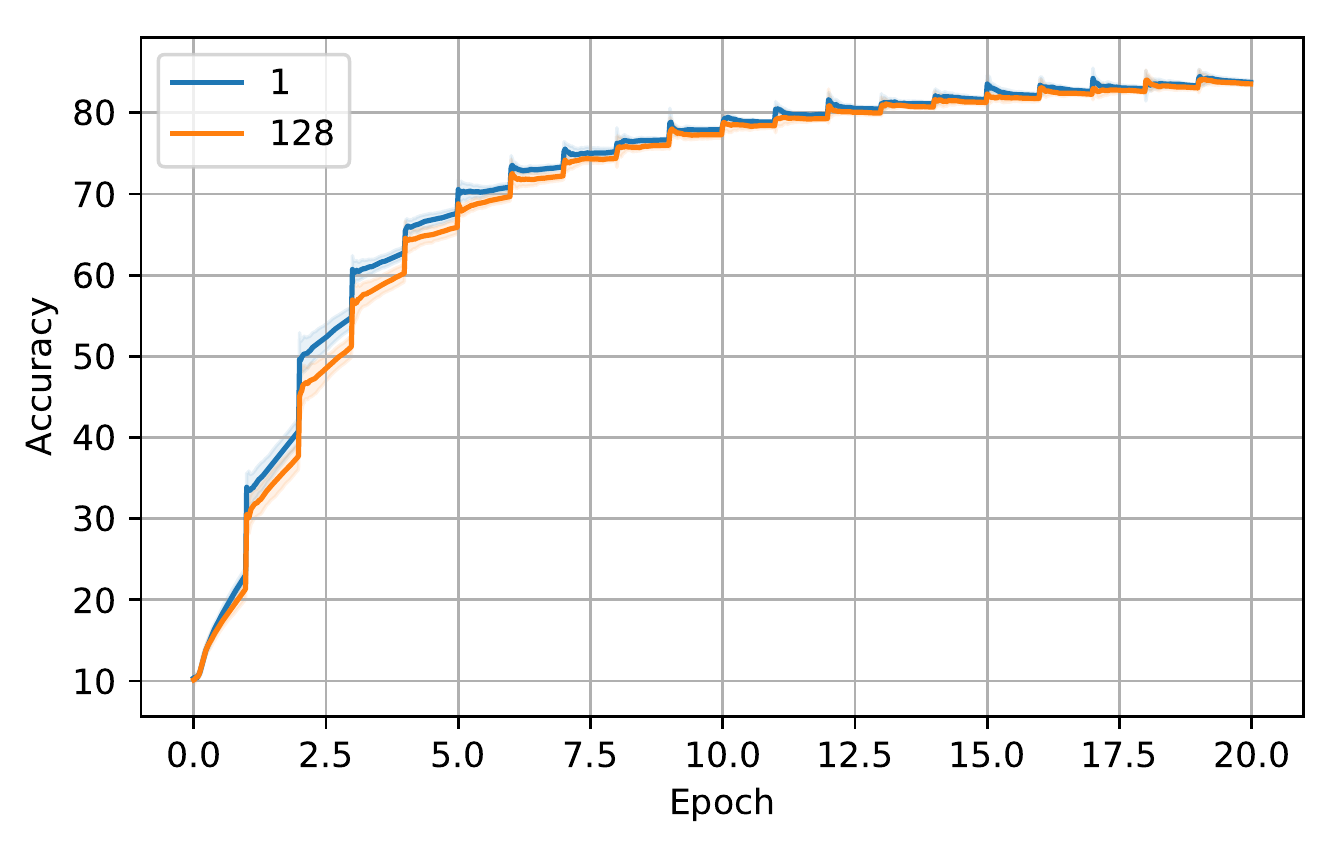}
         \caption{Training accuracy.}
         \label{fig:hp_scale_verif_t}
     \end{subfigure}
     \hfill
     \begin{subfigure}[b]{0.48\linewidth}
         \centering
         \includegraphics[width=\linewidth,height=4.7cm]{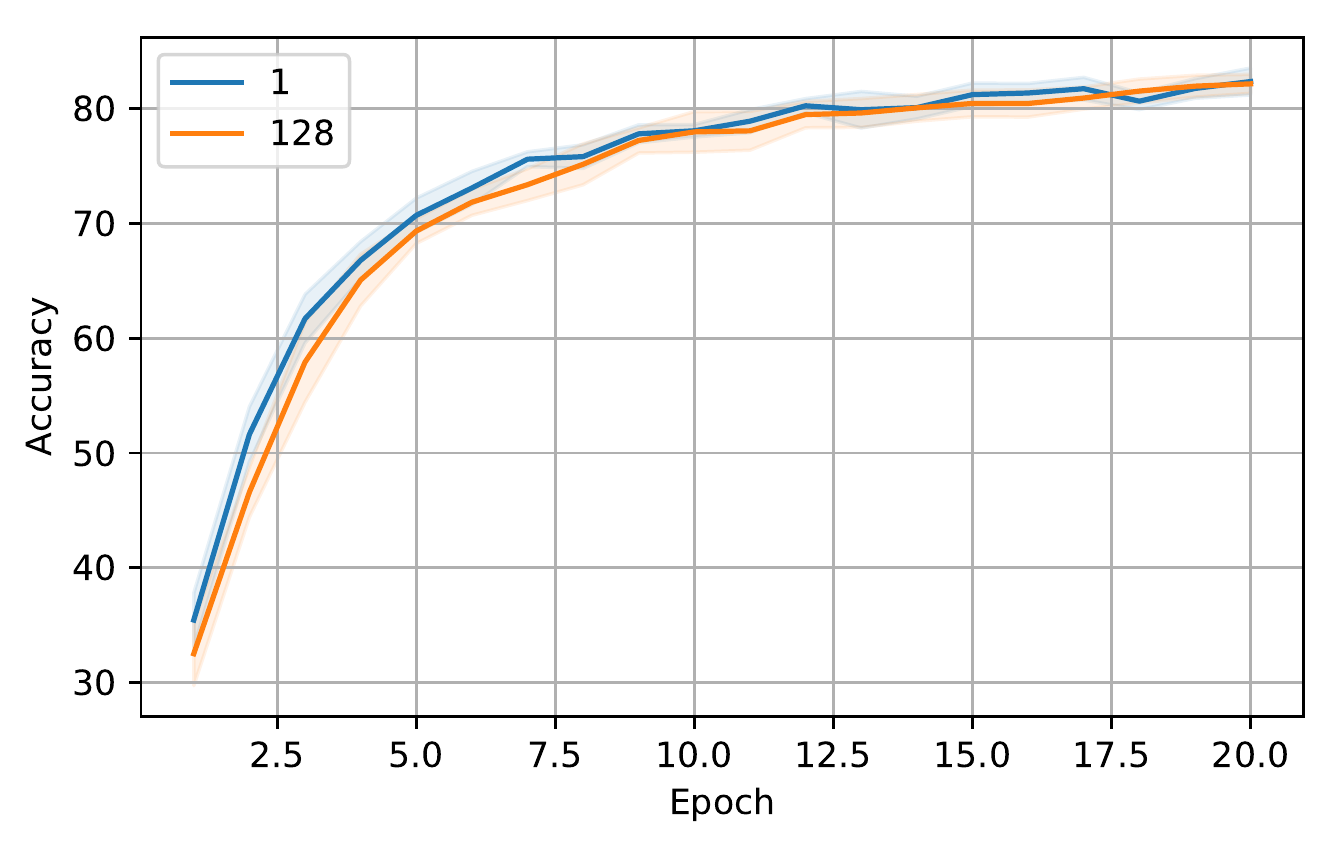}
         \caption{Validation accuracy.}
         \label{fig:hp_scale_verif_v}
     \end{subfigure}
        \caption{Hyperparameter comparison using CIFAR-10 VGG11. Showing mean (shading is standard deviation) of ten runs.}
        \label{fig:hp_scale_verif}
\end{figure*}

\section{Small Batch Size Training}
\label{sec:smallbt}

We define the per-worker batch size to be the number of samples that each pipeline stage processes at a time and the update-size to be the number of samples that contribute to the gradient in each update. We set both of these to one in our experiments. Alternatively the update-size could be set to match some reference for which known hyperparameters exist but this is outside the scope of this work.

Since the optimal learning rate and momentum depend on the update size $N$, we scale the values used by the SGDM reference according to \citeauthor{Chiley2019OnlineNF}~\citeyearpar{Chiley2019OnlineNF}. This corresponds to scaling the effective learning rate linearly with the update size while scaling the momentum such that the decay per sample is the same. 
This allows for a fair comparison of techniques even though different update sizes are used.
The scaling rules are:
\begin{align}
    m = m_{r}^{N/N_{r}} ,\qquad \eta = \frac{(1-m) N}{(1-m_{r}) N_r} \eta_r \label{eq:hpscale}
\end{align}
where $\eta_r$, $m_r$ and $N_r$ are the reference learning rate, momentum coefficient and update size and $\eta$, $m$ and $N$ are the new values (we use $N=1$).

Figures~\ref{fig:hp_scale_verif_t}~and~\ref{fig:hp_scale_verif_v} shows that the hyperparameters produced using these scaling rule result in batch size 1 training curves similar to the reference when training VGG11 on the CIFAR-10 dataset.
A similar result can be seen for ResNet training in Figure~\ref{fig:c10r20bsizeacc}.


\section{Inconsistent Weights vs Stale Gradients}
\label{sec:InconsistencyVsDelay}

In Pipelined Backpropagation gradients are delayed and computed with inconsistent weights. This can lead to accuracy degradation and instability.
In this section we investigate the relative importance of the effects.
We do this by comparing training with delayed gradients using either inconsistent or consistent weights.
In Appendix~\ref{pytorch_del_grad_sim} we describe how we can simulate this in PyTorch~\cite{paszke2017automatic} without using Pipelined Backpropagation.

\begin{figure}
    \centering
    \includegraphics[width=\linewidth]{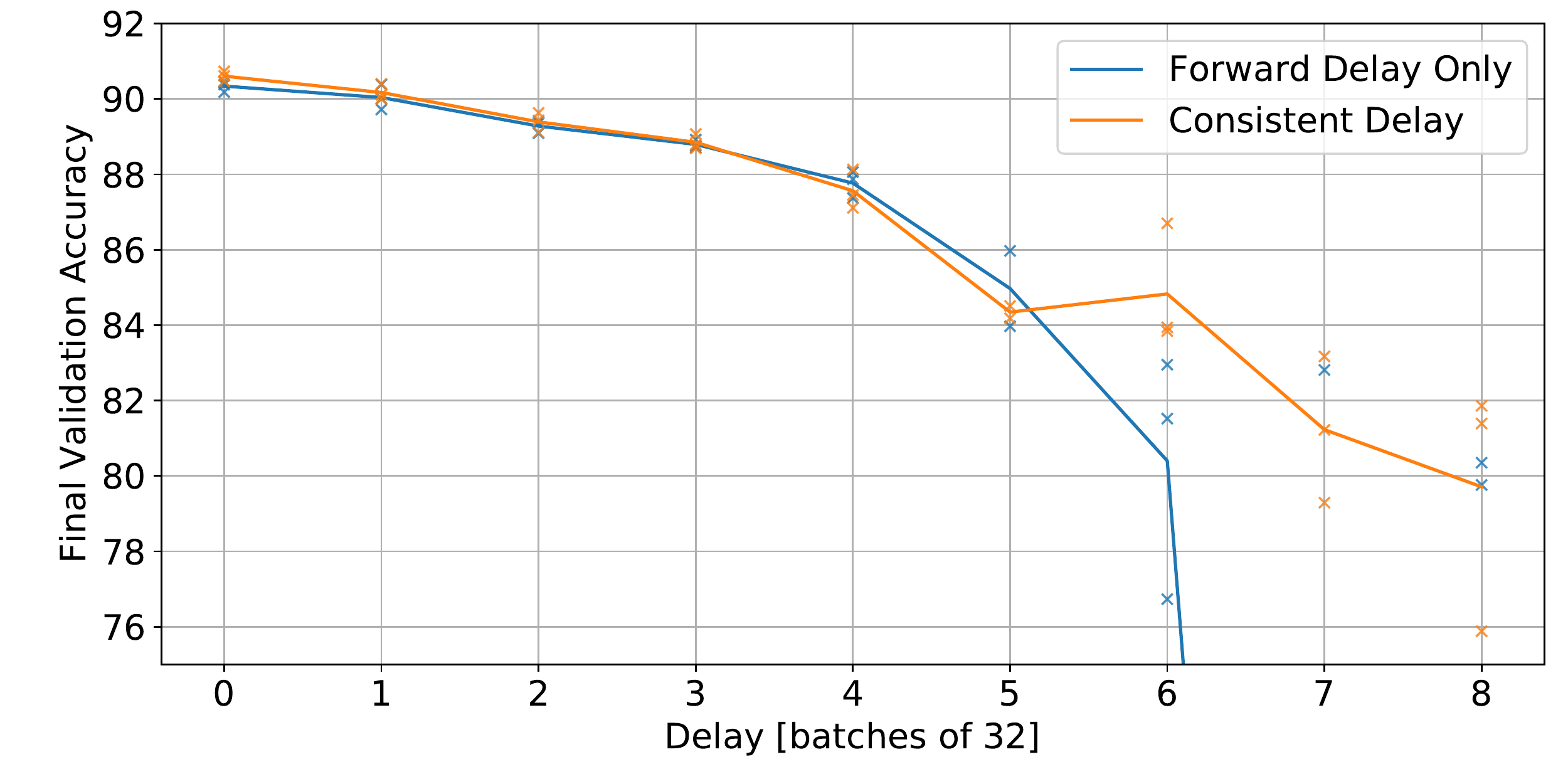}
    \caption{
    The effect of weight inconsistency on the final validation accuracy of CIFAR-10 ResNet-20 (with Group Normalization) for different delays. 
    \textbf{Consistent Delay} uses the same old version of the weights for both the forward and backward passes. This produces delayed gradients.
    \textbf{Forward Delay Only} uses old versions of the weights on the forward pass and current weights on the backwards pass, resulting in weight inconsistency.
    Delayed gradients result in a loss of final accuracy. Adding weight inconsistency only incurs additional degradation for large delays.
    }
    \label{fig:IncoVsDel}
\end{figure}

Figure~\ref{fig:IncoVsDel} shows the effects of delay on the final accuracy of CIFAR-10 ResNet-20 training with or without inconsistent weights.
As can be seen, even modest delays affect the final accuracy of training.
Weight inconsistency does not cause an additional loss of accuracy for small delays but causes a rapid loss of accuracy beyond a certain delay.
This transition point where weight inconsistency starts to affect training will depend on the dataset and architecture.
\citeauthor{Harlap2018PipeDreamFA}~\citeyearpar{Harlap2018PipeDreamFA} and \citeauthor{Chen2018EfficientAR}~\citeyearpar{Chen2018EfficientAR} make opposing claims about the effect of weight inconsistency.
\citeauthor{Harlap2018PipeDreamFA}~\citeyearpar{Harlap2018PipeDreamFA} introduce Weight Stashing to fix weight inconsistency and claim its use is necessary for convergence. 
\citeauthor{Chen2018EfficientAR}~\citeyearpar{Chen2018EfficientAR} show that Weight Stashing has no effect on training in their experiments so it should not be used to avoid memory overhead. 
Our results suggest that the effects of weight inconsistency depend on the magnitude of delays reconciling the two claims.

We also investigate the effect of weight inconsistency in our fine-grained Pipelined Backpropagation setup.
Table~\ref{tab:c10accMeanRoundrtr} compares PB training with and without Weight Stashing.
The results suggest that Weight Stashing is not beneficial in our setup so we do not use it in other experiments.
This indicates that weight inconsistency is likely not an issue and the accuracy losses of PB primarily stem from the gradient delay.
As mentioned in the conclusion, the small batch sizes we use combined with the hyperparameter scaling may reduce the effects of the delay.
For larger batch sizes weight inconsistency may be a bigger issue.


\section{Forms of Weight Prediction}
\label{apdx:wp_forms}

The goal of weight prediction is to approximate future weights to combat gradient delay and weight inconsistency.
Linear Weight Prediction (LWP) gives a general form for predicting the network state $T$ steps into the future by using the velocity. 
In Pipelined Backpropagation the delay varies for different stages. 
By default (LWP\textsubscript{D}) we set $T$ equal to the delay for every stage (see red arrows in Figure~\ref{fig:methods_overhead} left).
Other works have proposed related forms of weight prediction.

LWP is closely related to the weight prediction proposed in SpecTrain~\cite{Chen2018EfficientAR}.
SpecTrain extends the prediction horizon such that all stages predict to the same time step.
This form of time synchronization is first described by \citeauthor{Harlap2018PipeDreamFA}~\citeyearpar{Harlap2018PipeDreamFA} as Vertical Sync. 
The forward prediction horizon is depicted in green in Figure~\ref{fig:methods_overhead} left. 
With the extended prediction horizon, SpecTrain must also predicts weights on the backwards pass to address inconsistency. The prediction horizon for the backward pass weights is depicted in blue in Figure~\ref{fig:methods_overhead} left. 
This can be seen as using a stage dependent extended  prediction horizon (Appendix~\ref{apdx:wp2}). 

Discrepancy correction~\cite{yang2019pipemare} can be seen as a form of weight prediction. Whereas LWP and SpecTrain predict weights into the future to mitigate for gradient delay and weight inconsistency, PipeMare approximates the weights used on the forward pass during the backward pass. This can only deal with weight inconsistency, but potentially provides a more accurate prediction. 
Discrepancy correction uses a separate exponential tracker for their prediction. LWP uses the optimizer velocity directly.
In Appendix~\ref{sec:InconsistencyVsDelay} we show that weight inconsistency is not a significant issue in our setting so we primarily focus on mitigating the effects of gradient delay.

DANA~\cite{hakimi2019taming} is another variant of weight prediction that has been used in the ASGD setting but is not directly applicable to Pipelined Backpropagation.


\section{Mitigation Overhead}
\label{sec:methods_overhead}

\begin{figure}
    \begin{subfigure}[b]{.49\linewidth}
        \centering
        \includegraphics[width=\linewidth]{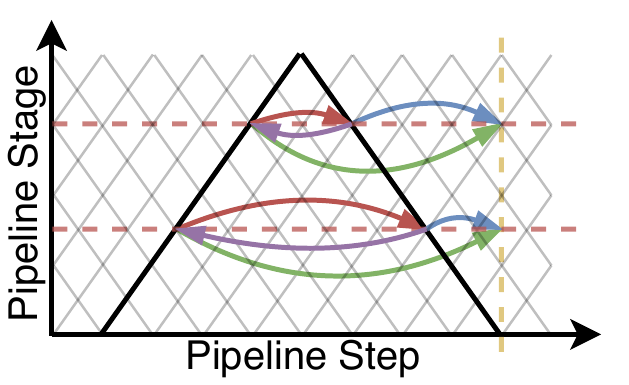}
    \end{subfigure}
    \hfill
    \begin{subfigure}[b]{.49\linewidth}
        \centering
        \includegraphics[width=\linewidth]{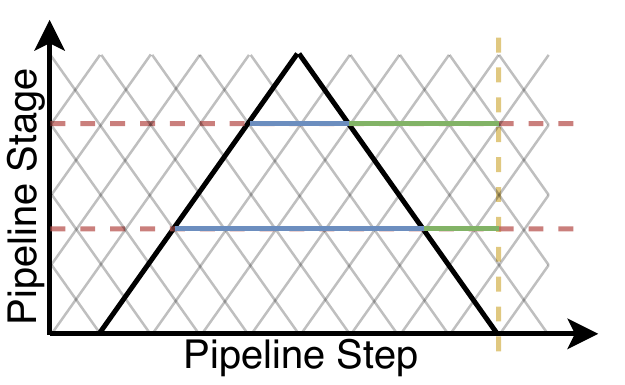}
    \end{subfigure}
    \caption{Compute and Memory Overhead of PB Mitigations. Two of the stages are highlighted in red dashed lines and the yellow dashed line depicts a vertical synchronization boundary. \textbf{Left}: Shows the forward and backward prediction horizons of SpecTrain in green and blue respectively. The forward prediction horizon of LWP$_\mathrm{D}$ is shown in red and the backward prediction horizon of Pipemare's discrepancy correction is shown in purple. \textbf{Right}: show the buffer lengths for Pipedreams Weight Stashing in blue and the gradient buffers of Vertical Sync in green.}
    \label{fig:methods_overhead}
\end{figure}

Figure~\ref{fig:methods_overhead} highlights two network stages and their associated overheads for different mitigation methods.
Only two stages are highlighted but these overheads exist for every stage in the network.
The two stages are highlighted with red dashed lines.
The yellow dashed line shows a vertical synchronization boundary.
Figure~\ref{fig:methods_overhead} left depicts the forward and backward weight predictions of SpecTrain in green and blue, respectively.
SpecTrain has the memory overhead of storing the forward and backward state as well as the compute overhead of predicting those states.
The forward weight prediction of LWP is shown in red which has exactly half the overhead of SpecTrain.
Pipemare's discrepancy correction technique is a form of backward weight prediction (shown in purple) and has the compute and memory overhead of storing and predicting the forward pass network state for use in the backwards pass; they also introduce a state transition tracker to make this prediction.
Their proposed warmup epochs would also incur the fill and drain overhead of pipelined training.
Figure~\ref{fig:methods_overhead} right shows the weight and gradient tensor buffers required for Pipedream's Weight Stashing and Vertical Sync in blue and green respectively.
While Pipedream has by far the largest memory overhead, their methods which have no compute overhead.
SC and Gradient Shrinking simply modify the weight update equations and have no memory or compute overhead.


\section{State Transition Equations}
\label{apdx:state_transition}
In order to analyze and compare our methods, we view the optimization as a dynamical system in terms of its state transition equation. A similar approach is used in \cite{o2015adaptive,goh2017why,giladi2019stability}. We assume that $\bar{L}(w_t)$ is the underlying loss function we are trying to minimize where $w_t$ are the weights at time $t$. For neural networks, $\bar{L}$ could be the mean training loss, the expected loss over all training samples. We assume that for a given sample or time step, the gradient with respect to the weights is $\nabla \bar{L}(w_t) + R$ where $R=R(w_t)$ is a random variable. 
The expectation of $R$ (over all samples) is assumed to be zero.

We are interested in comparing the dynamics of delayed SGDM, weight prediction, Spike Compensation and the combined mitigation. These can all be seen as special cases of the combined mitigation given in Section~\ref{sec:combined_mitigation} for the appropriate choice of $a$, $b$ and $T$. The velocity form of the combined mitigation, LWP$^{\mathrm{v}}$+SC, results in a complicated state transition equation which can not be easily analyzed without further simplifications. The velocity form can be approximated with the weight difference form, LWP$^{\mathrm{w}}$+SC. This form is simple to analyze so we use it for the rest of the analysis.

We analyze the systems in expectation and do not try to estimate the variance. Let $\bar{w}_t$ and $\bar{v}_t$ be the expected weights and velocity at time $t$. We can then write the expected state update for the combined mitigation at time $t$ in terms of previous expected values as:
\begin{align}
    \bar{v}_{t+1} &= \E[m \bar{v}_{t} + g_t] \nonumber \\
    &= m \bar{v}_{t} + \bar{g}_t \label{eq:exp_v_update} \\
    \bar{w}_{t+1} &= \E[\bar{w}_t - \eta \cdot \left( a \bar{v}_{t+1} + b g_t \right)] \nonumber \\
    &= \bar{w}_t - \eta \cdot \left( a \bar{v}_{t+1} + b \bar{g}_t \right) \label{eq:exp_w_update}
\end{align}
where $a$, $b$ are the coefficients for General Spike Compensation and $\bar{g}_t := \E[g_t]$ is the expected gradient arriving at time $t$. This gradient is calculated using weight prediction with horizon $T$ from weights delayed by $D$ time steps:
\begin{align}
    \bar{g}_t &= \E[\nabla \bar{L} \left( \bar{w}_{t-D} + T\cdot(\bar{w}_{t-D} - \bar{w}_{t-D-1}) \right) + R] \nonumber \\
    &= \nabla \bar{L} \left( \bar{w}_{t-D} + T\cdot(\bar{w}_{t-D} - \bar{w}_{t-D-1}) \right) \label{eq:exp_g}
\end{align}

We can isolate $\bar{v}_{t+1}$ from equation~\eqnref{eq:exp_w_update}:
\begin{equation}
    \bar{v}_{t+1} = \frac{-1}{\eta a}\left( \bar{w}_{t+1} - \bar{w}_{t} \right) - \frac{b}{a}\bar{g}_t
\end{equation}
Shifting the time index we obtain an expression for $\bar{v}_t$ which we can insert into equation~\eqnref{eq:exp_v_update}:
\begin{equation}
    \bar{v}_{t+1} = \frac{-m}{\eta a}\left( \bar{w}_t - \bar{w}_{t-1} \right) - \frac{bm}{a}\bar{g}_{t-1} + \bar{g}_t \label{eq:v_from_w}
\end{equation}
Combining equations~\eqnref{eq:exp_w_update}, \eqnref{eq:exp_g} and  \eqnref{eq:v_from_w} we obtain a state transition equation in terms of the expected weights without the velocity:
\begin{align}
    \bar{w}_{t+1} &= (1+m) \bar{w}_t - m \bar{w}_{t-1} \\
    &-\eta \cdot (a+b) \nabla\bar{L}\left((T+1)\bar{w}_{t-D} - T \bar{w}_{t-D-1}) \right) \nonumber\\
    &+\eta m b \nabla\bar{L}\left((T+1)\bar{w}_{t-D-1}- T \bar{w}_{t-D-2}) \right) \nonumber
\end{align}

By inserting appropriate values for $T$, $a$ and $b$ we can obtain the state transition equations for General Spike Compensation (GSC, $T=0$), linear weight prediction (LWP, $a=1, b=0$) and SGDM with delay ($a=1, b=0, T=0$):
\begin{align}
    \text{SGDM:}\; \bar{w}_{t+1} = &\; (1 + m) \bar{w}_t - m \bar{w}_{t-1} \\ 
        & - \eta \nabla\bar{L}(\bar{w}_{t-D}) \nonumber \\
    \text{GSC:}\; \bar{w}_{t+1} = &\; (1+m) \bar{w}_t  - m \bar{w}_{t-1} \label{eq:gscst} \\
        & - \eta \cdot (a+b) \nabla\bar{L}(\bar{w}_{t-D}) \nonumber \\
        & + \eta m b \nabla\bar{L}(\bar{w}_{t-D-1})  \nonumber \\
    \text{LWP:}\; \bar{w}_{t+1} = & \;(1+m) \bar{w}_t - m \bar{w}_{t-1} \label{eq:lwpst} \\ 
        & - \eta \nabla\bar{L}\left((T+1) \bar{w}_{t-D} - T \bar{w}_{t-D-1}\right) \nonumber
\end{align}
We note that unlike state transition equation of SGDM the equations for LWP and GSC both contain $\bar{w}_{t-D-1}$. This means that the mitigation methods generally do not correspond to a simple change in the hyperparameter values of SGDM. Similarly, the combination of GSC and LWP has an additional $\bar{w}_{t-D-2}$ term and thus does not simply correspond to a different setting of $a$, $b$ or $T$ for either method.

\subsection{Comparing LWP and GSC}
\label{apdx:lwp_vs_gsc}
The equations for LWP and GSC contain the same weight terms which could indicate that they operate in similar ways. If the gradient is well approximated as a linear function on the line segment:
\begin{equation*}
    \{\bar{w}_{t-D-1} + \alpha (T+1) (\bar{w}_{t-D}-\bar{w}_{t-D-1})\;|\;\alpha \in [0;1] \}
\end{equation*}
we have:
\begin{equation}
\begin{split}
    & \nabla\bar{L}\left((T+1) \bar{w}_{t-D} -T \bar{w}_{t-D-1}\right) \\
    & \quad \approx (T+1) \nabla\bar{L}(\bar{w}_{t-D}) - T \nabla\bar{L}(\bar{w}_{t-D-1}) \label{eq:lina}
\end{split}
\end{equation}
In this case GSC and LWP are equivalent for the same learning rate and momentum if:
\begin{align}
    a + b &= 1 + T \label{eq:wp_sc_eq_cond1} \\
    mb &= T \label{eq:wp_sc_eq_cond2}
\end{align}
When the approximation in equation~\eqnref{eq:lina} holds, LWP is equivalent to our default choice of $a$ and $b$ (see equation~\eqnref{eq:scd}) if:
\begin{equation}
    T = m \frac{1-m^D}{1-m}
\end{equation}
This is equivalent to assuming zero future gradient over the prediction horizon in equation~\eqnref{eq:wpvariants} instead of a constant velocity. GSC is equivalent to LWP with horizon $T$ for the same learning rate if the approximation in \eqnref{eq:lina} holds and:
\begin{align}
    a = 1 - \frac{1-m}{m} T,\qquad b = \frac{T}{m}
\end{align}

This shows that LWP and GSC are closely related. Both methods compensate for a delay but at different points in time. Weight prediction changes how the gradient is computed, Spike Compensation changes how it is applied. Each method has its advantages. Spike Compensation has minimal overhead and doesn't require an estimate of the delay ahead of time. Weight prediction might introduce memory overhead by adding a new copy of the weights (depending on the implementation and hardware), but may help reduce weight inconsistency. The combination of the two methods can be useful in cases where we want to overcompensate for the delay. A similar effect can be achieved with either method by changing the horizon but their combination offers increased weight consistency without requiring an additional weight prediction on the backwards pass.


\section{Extended Weight Prediction Horizons}
\label{apdx:wp2}

\begin{figure}[htb]
    \centering
    \includegraphics[width=\linewidth]{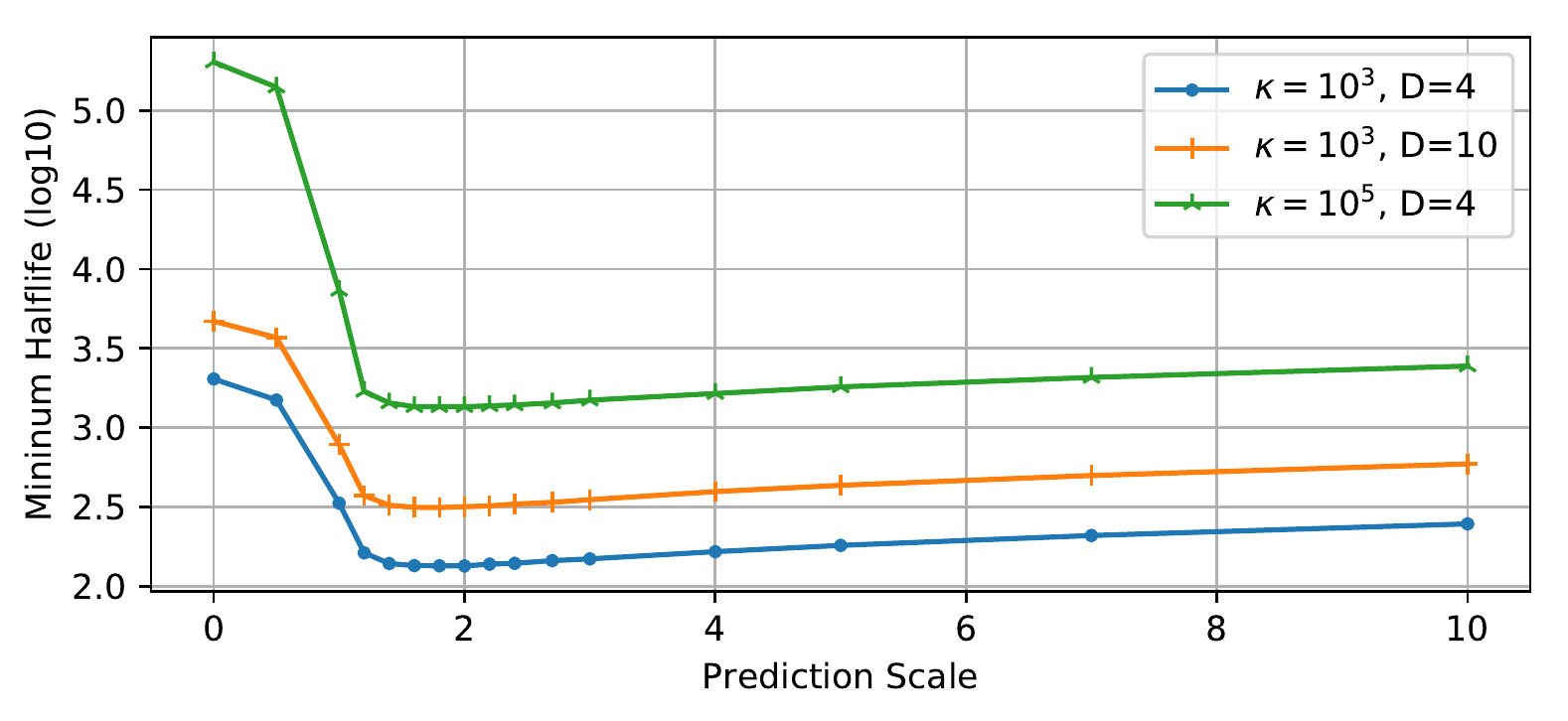}
    \caption{The convergence speed for a convex quadratic with different condition numbers ($\kappa$) and delays ($D$). A weight prediction with horizon $T=\alpha D$ is used where $\alpha$ is the prediction scale shown on the horizontal axis.}
    \label{fig:extended_wp_horizon_cq}
\end{figure}

In Section~\ref{sec:cq_analysis} we discuss how overcompensating for delays can help improve convergence speed. One way to do this is to predict weights more than $D$ (the delay) steps into the future with linear weight prediction. Figure~\ref{fig:extended_wp_horizon_cq} shows the effect of scaling the weight prediction horizon on the convergence rate when optimizing a convex quadratic. We see that horizon lengths of around $T=2D$ seem to give the best results.

\begin{figure}[htb]
    \centering
    \includegraphics[width=\linewidth]{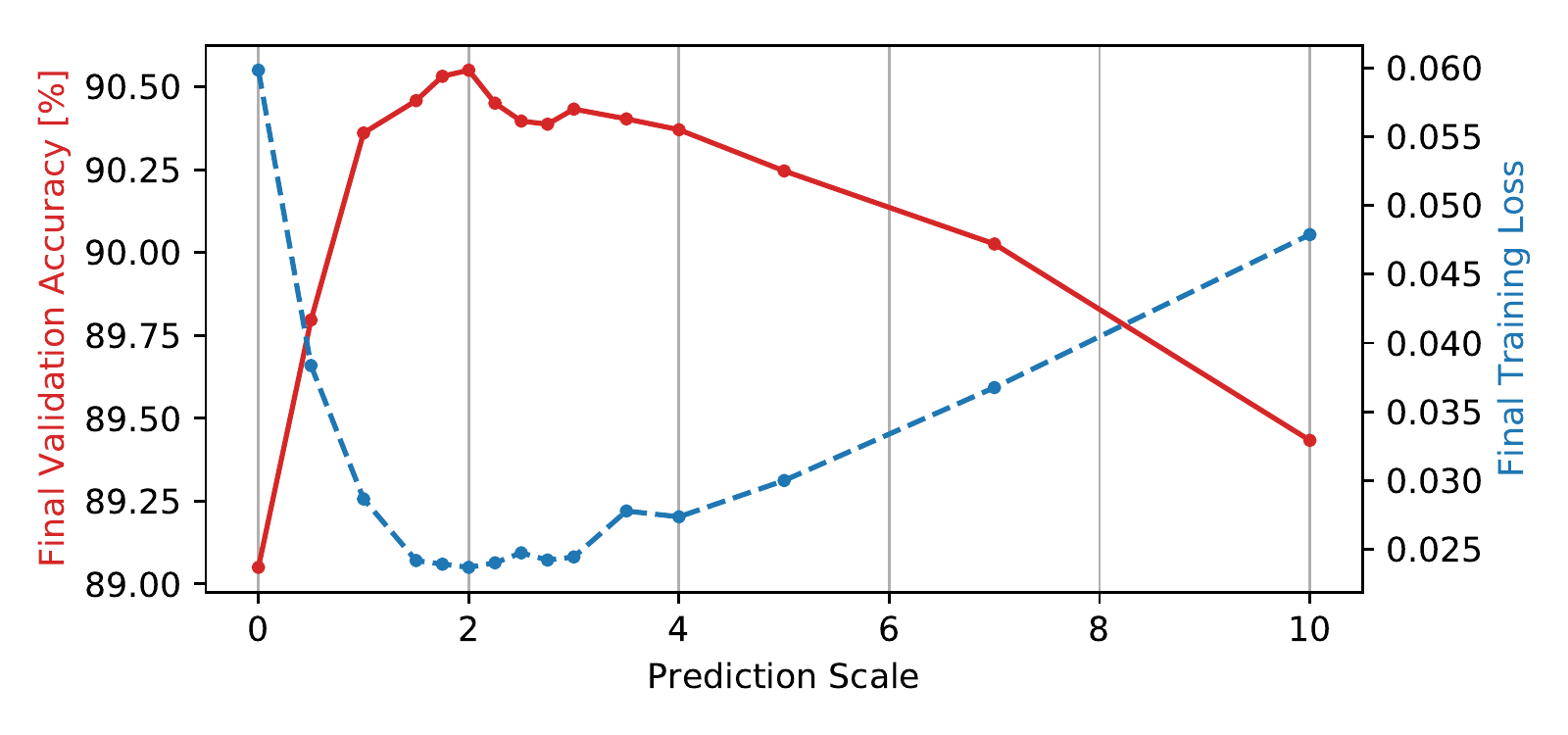}
    \caption{The effects of different weight prediction horizons on the final loss and accuracy when training ResNet-20 on CIFAR-10. A prediction scale of $\alpha$ scales the horizon to be $T=\alpha D$ where $D=4$ is the delay. The delay is the same for all layers and consistent weights are used. Each point is the mean of multiple runs, 25 for $1.75\le\alpha\le2.5$, and 10 for other $\alpha$ values.
    }
    \label{fig:extended_wp_horizon_rn}
\end{figure}

We repeated this experiment for ResNet-20 (with group normalization) trained on CIFAR-10 using the simplified delay setup described in Appendix \ref{pytorch_del_grad_sim}. We used a delay $D=4$ for all layers with consistent weights and a batch size of 32 for a total delay of 128 samples (which is in the range of many of our PB experiments). The learning rate and momentum were scaled according to \eqnref{eq:hpscale} using the default reference values referenced in the experiments section. The results can be seen in Figure~\ref{fig:extended_wp_horizon_rn}. We can see that the training loss curve looks somewhat similar to the convergence speed for the convex quadratic, with the lowest loss obtained for $T\approx 2D$. The validation accuracy also peaks for $T\approx 2D$.

\begin{table*}
    \caption{CIFAR-10 validation accuracy (mean$\pm$std.dev of five runs) for ResNet (RN) and VGG training with overcompensation.}
    \label{tab:c10accLWP2D}
    \centering
    \vskip 0.15in
    \small
    \sc
    \begin{tabular}{l|cc|cc|cc}
        \toprule
        Network & SGDM                    & PB                      & PB+LWP$_\mathrm{D}$          & PB+LWP$_\mathrm{2D}$            & PB+SC$_\mathrm{D}$      & PB+SC$_\mathrm{2D}$     \\ \midrule
        VGG11   & \textbf{91.16}$\pm$0.19 & 90.83$\pm$0.20          & 91.05$\pm$0.11               & \textbf{91.27}$\pm$0.14         & \textbf{91.08}$\pm$0.19 & \textbf{91.03}$\pm$0.22 \\
        VGG13   & \textbf{92.57}$\pm$0.15 & \textbf{92.59}$\pm$0.15 & \textbf{92.51}$\pm$0.11      & \textbf{92.57}$\pm$0.21         & 92.38$\pm$0.27          & \textbf{92.60}$\pm$0.17 \\
        VGG16   & \textbf{92.24}$\pm$0.19 & 92.06$\pm$0.21          & \textbf{92.22}$\pm$0.24      & \textbf{92.28}$\pm$0.18         & \textbf{92.45}$\pm$0.30 & \textbf{92.42}$\pm$0.21 \\ \midrule
        RN20    & \textbf{90.63}$\pm$0.31 & 90.44$\pm$0.24          & 90.68$\pm$0.30               & \textbf{91.05}$\pm$0.10         & \textbf{90.80}$\pm$0.29 & \textbf{90.95}$\pm$0.40 \\
        RN32    & \textbf{91.68}$\pm$0.23 & 91.46$\pm$0.09          & 91.66$\pm$0.10               & \textbf{91.98}$\pm$0.22         & 91.55$\pm$0.14          & \textbf{91.96}$\pm$0.24 \\
        RN44    & \textbf{92.19}$\pm$0.14 & 91.71$\pm$0.25          & 92.00$\pm$0.14               & \textbf{92.29}$\pm$0.09         & \textbf{92.13}$\pm$0.16 & \textbf{92.21}$\pm$0.21 \\
        RN56    & \textbf{92.39}$\pm$0.20 & 91.89$\pm$0.40          & 92.31$\pm$0.14               & \textbf{92.41}$\pm$0.17         & 92.33$\pm$0.16          & \textbf{92.68}$\pm$0.23 \\
        RN110   & \textbf{92.77}$\pm$0.22 & 91.81$\pm$0.15          & \textbf{92.76}$\pm$0.05      & 71.83$\pm$36.91\footnotemark[2] & \textbf{92.28}$\pm$0.29 & \textbf{92.35}$\pm$0.85 \\
        \bottomrule
    \end{tabular}
    \vskip -0.1in
\end{table*}

\begin{figure*}[ht]
     \centering
     \begin{subfigure}[t]{0.48\linewidth}
         \centering
         \includegraphics[width=\linewidth]{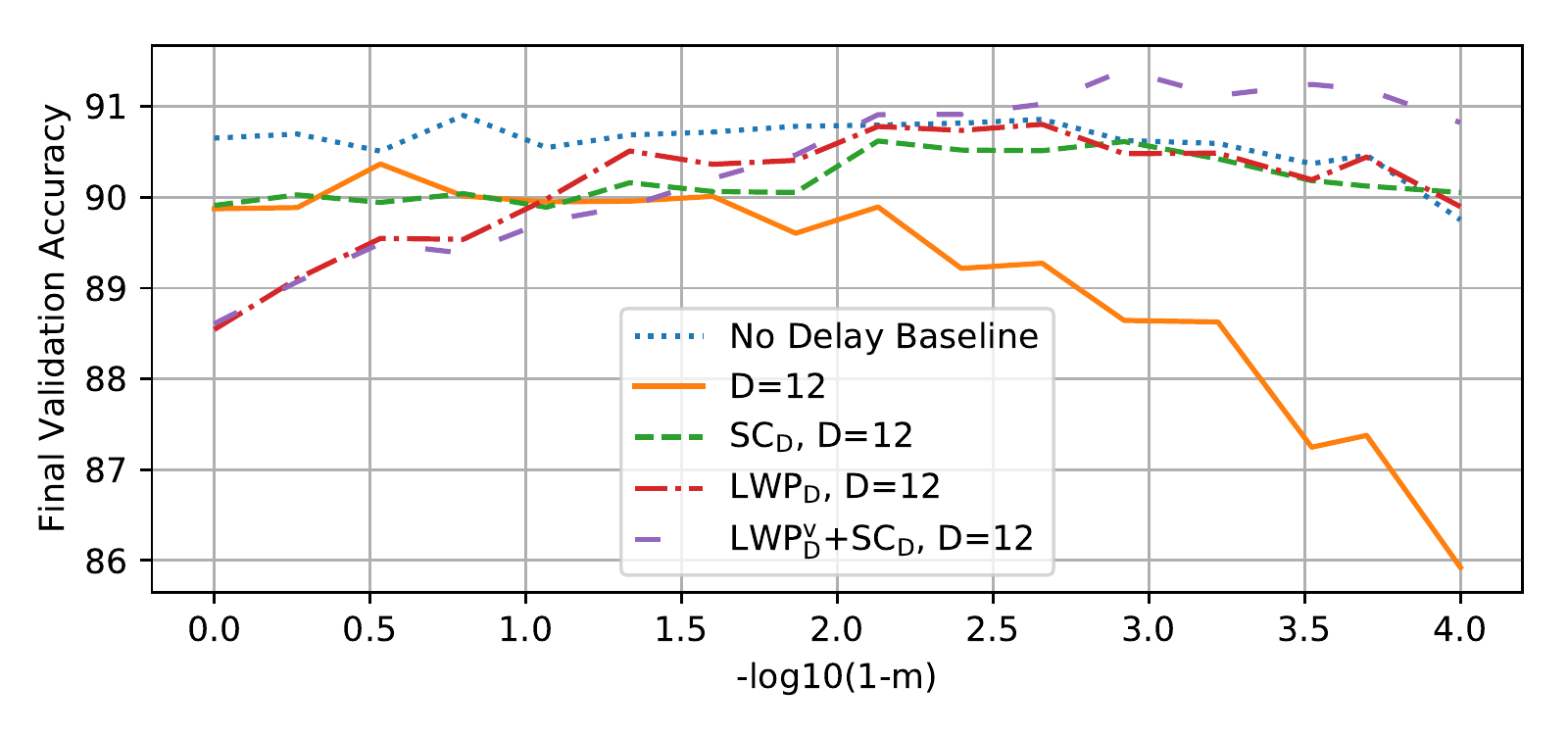}
         \caption{Using consistent weights.}
         \label{fig:mom_sweep_rn20_ws}
     \end{subfigure}
     \hfill
     \begin{subfigure}[t]{0.48\linewidth}
         \centering
         \includegraphics[width=\linewidth]{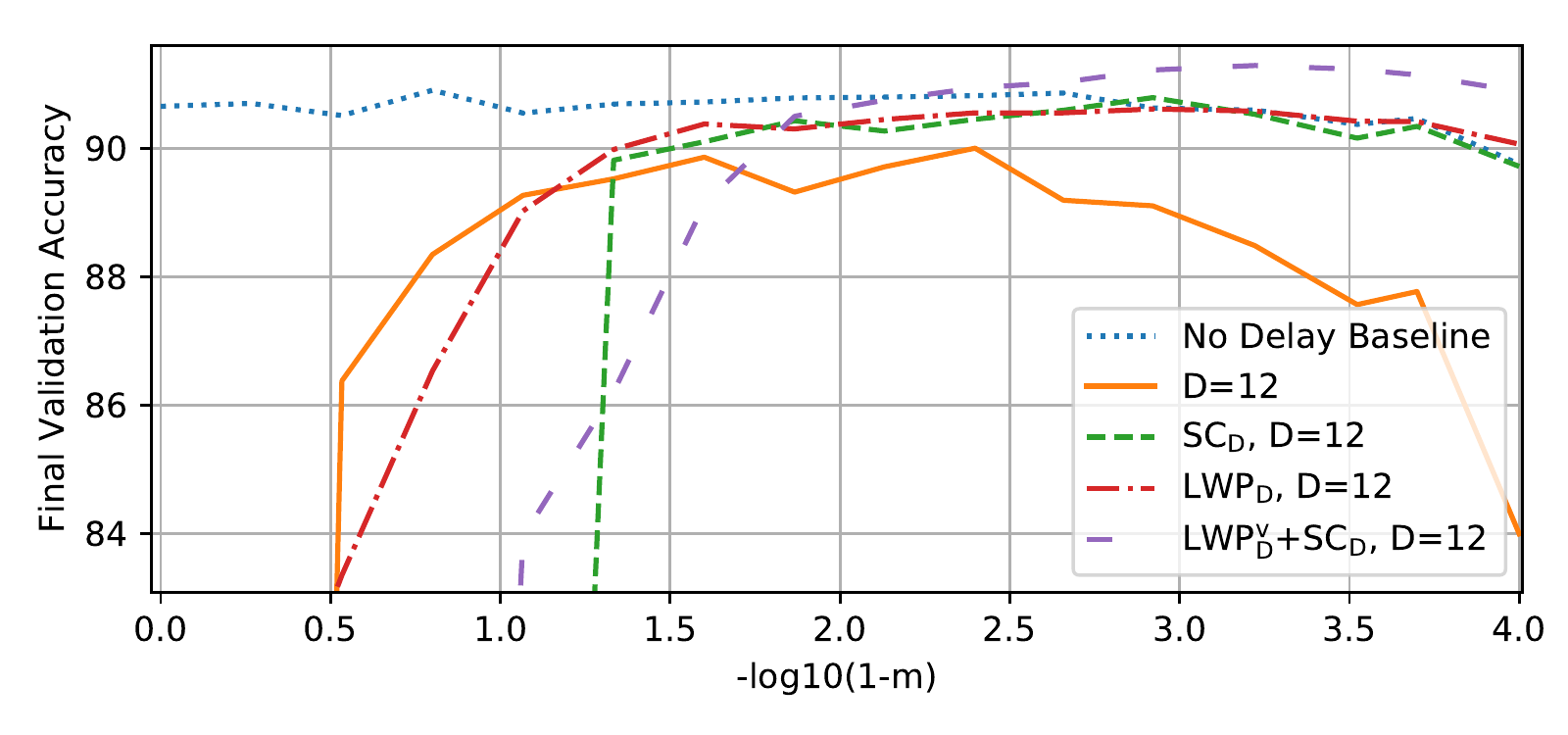}
         \caption{Using inconsistent weights.}
         \label{fig:mom_sweep_rn20_cp}
     \end{subfigure}
        \caption{
            Effect of momentum on CIFAR-10 ResNet-20 training with delay. Showing the mean of three runs (six for the no-delay case).
        }
        \label{fig:mom_sweep_rn20}
\end{figure*}

We also test this hypothesis in the Pipelined Backpropagation setting. We explore the use of weight prediction with a horizon which is double that of the delay (LWP$_\mathrm{2D}$). We also experiment with overcompensating for the delay by doubling the effect of Spike Compensation (SC$_\mathrm{2D}$ which replaces D with 2D in \eqnref{eq:scd}).
We observe that overcompensating can improve the final accuracy in most cases (Table~\ref{tab:c10accLWP2D}\footnote{In each row, in each column pair, the values within one standard error of the maximum accuracy are highlighted.}).
We note that in these networks weight inconsistency does not seem to be an issue (see Appendix~\ref{sec:InconsistencyVsDelay}). 
In cases where weight inconsistency is an issue, doubling the prediction horizon can reduce training stability. 
The same may apply to networks with large delays.
One such example may be training ResNet-110 on CIFAR-10 (Table~\ref{tab:c10accLWP2D}) where standard weight prediction outperforms methods which overcompensate for delay.


\section{Effects of Momentum Scaling}
\label{apdx:momentum_scaling_effect}

Throughout this work we heuristically scale the momentum and learning rate for small batch size training according to \eqnref{eq:hpscale}. This enables us to use Pipelined Backpropagation without further hyperparameter tuning for existing networks which is important for the practicality of PB training. These rules increase the momentum significantly compared to other heuristics which might keep it constant or lower it. In Section~\ref{sec:cq_analysis} we show that momentum loses some of its benefits with delays. However, our compensation methods, Spike Compensation and Linear Weight Prediction, likely benefit from high momentum. In this section we look at the effects of different momentum values, while keeping the total contribution from each gradient the same. We do this by selecting a specific value of $m$ in \eqnref{eq:hpscale} (ignoring the first expression) and then scaling the learning rate according to the second expression.

The experiments involve training ResNet-20 (with group normalization) on CIFAR-10 using the simplified delay setup described in Appendix \ref{pytorch_del_grad_sim}. We use a batch size of 8 and a delay of 12 for all layers for a total delay of 96 samples (which is in the range of many of our PB experiments). Figure~\ref{fig:mom_sweep_rn20_ws} shows this when consistent weights are used. We can see that for the baseline with no delay a wide range of momentum values can be used, including no momentum, but very large values cause accuracy loss. With delay, small values of momentum are better and the accuracy falls off relatively quickly for larger values. With our compensation methods the best accuracy is obtained for large momentum values. Spike Compensation has no effect for low (zero) momentum values and therefore matches the delayed baseline for small momentum values. Weight prediction for small momentum values tries to predict future weights based on recent gradients without sufficient smoothing and performs worse than the baseline. The combined mitigation exceeds the best results for the no-delay baseline for a range of large momentum values.

\begin{figure*}[ht]
     \centering
     \begin{subfigure}[b]{0.48\linewidth}
         \centering
         \includegraphics[width=\textwidth]{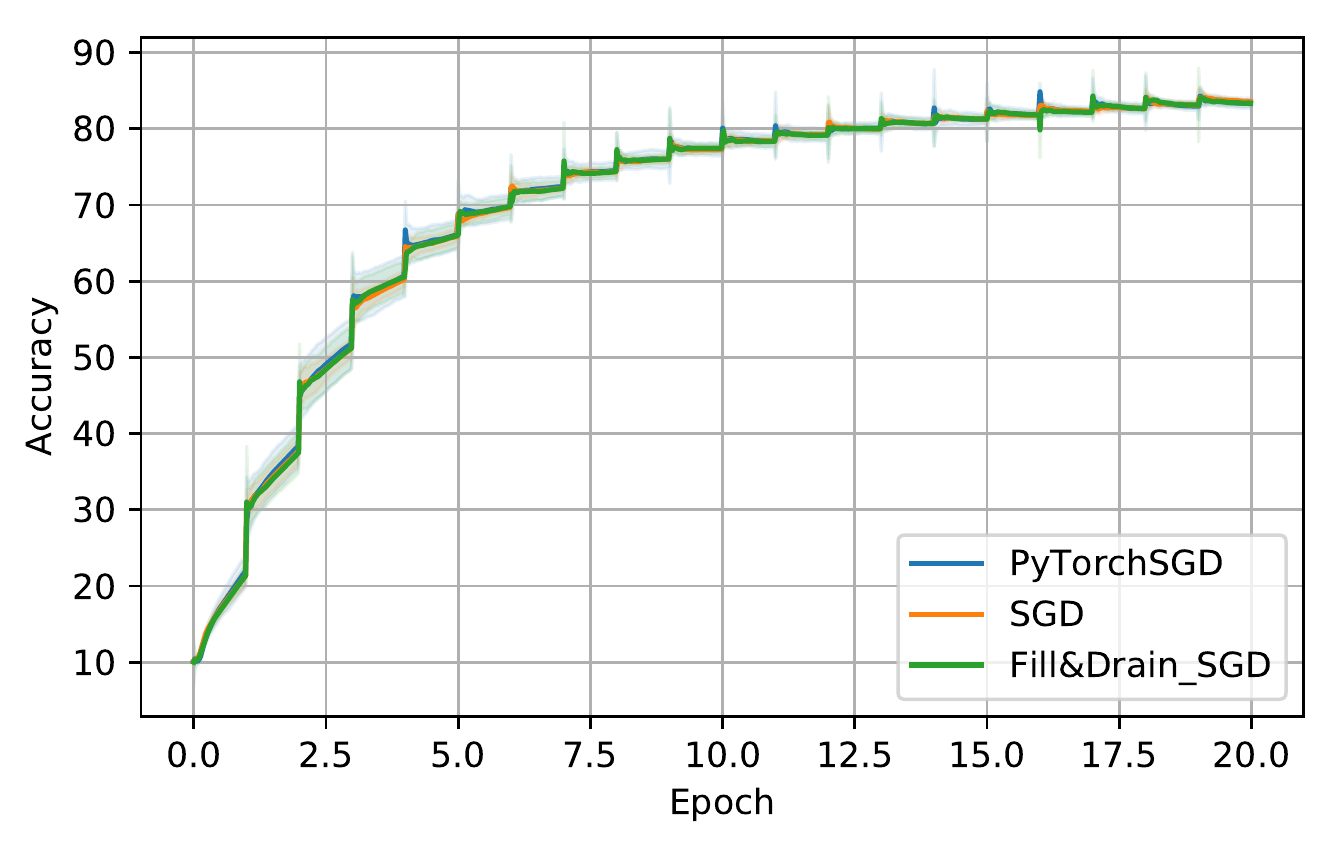}
         \caption{Training accuracy.}
         \label{fig:gprop_validation_t}
     \end{subfigure}
     \hfill
     \begin{subfigure}[b]{0.48\linewidth}
         \centering
         \includegraphics[width=\textwidth]{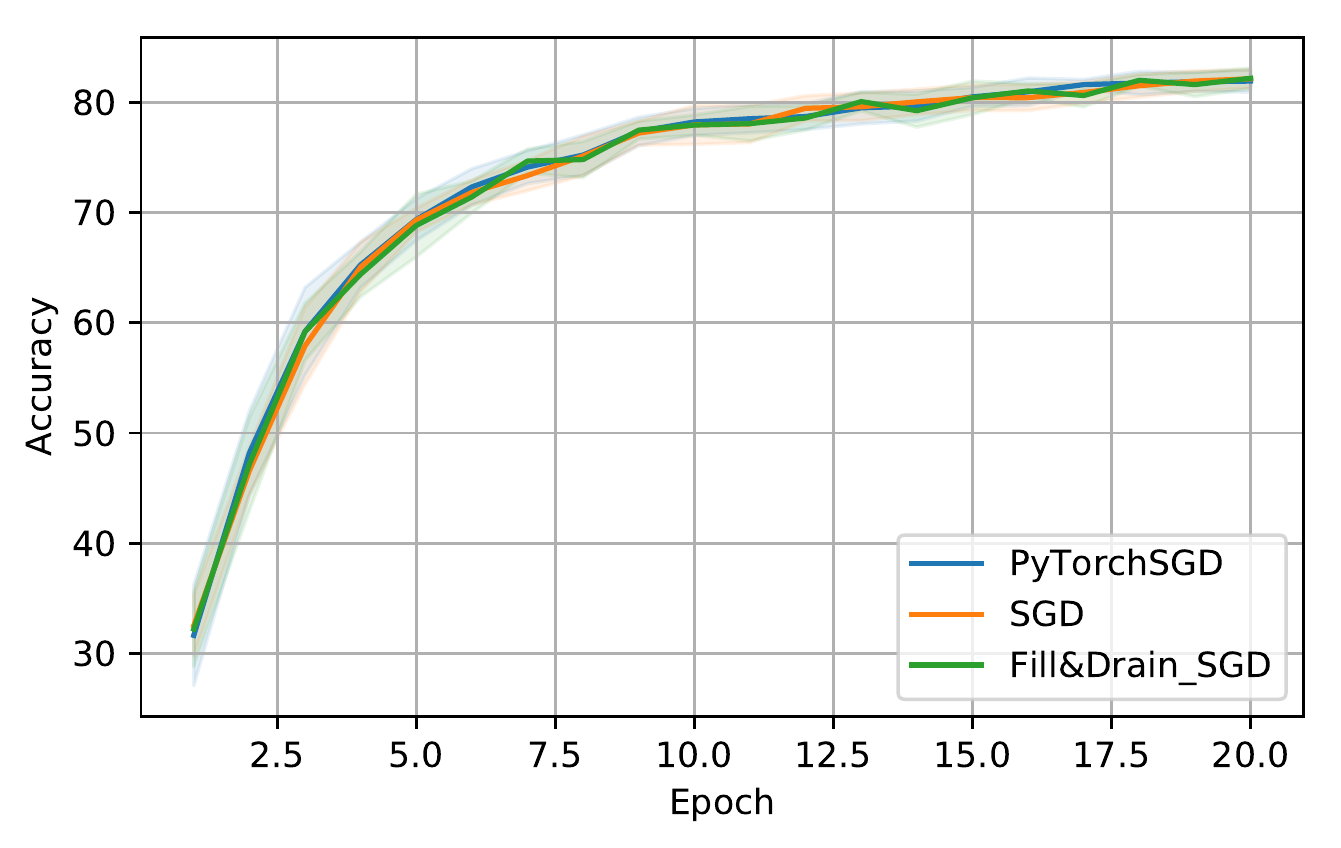}
         \caption{Validation accuracy.}
         \label{fig:gprop_validation_v}
     \end{subfigure}
        \caption{Validation of the GProp framework using CIFAR-10 VGG11. Showing mean (shading is standard deviation) of ten runs.}
        \label{fig:gprop_validation}
\end{figure*}

Figure~\ref{fig:mom_sweep_rn20_cp} shows the same experiment performed with inconsistent weights (using the most recent weights on the backwards pass instead of the delayed weights used on the forward pass). Most of the observations from the previous experiment hold in this case as well. The most notable difference is the poor performance of all methods when low momentum is used. This suggests that small momentum values adversely affect weight consistency. These runs do not use a tuned learning rate or a learning rate warmup which could likely help stabilize lower momentum values. Using our formulation of momentum causes a warmup in the step size while the velocity is building up. 
This effect could contribute to larger momentum values performing better.
Another factor may be the exponential smoothing of weight updates with momentum.
Without this, a couple of relatively large gradients could cause a large weight inconsistency for some time steps, potentially destabilize training.


\subsection{Tuning Momentum vs Spike Compensation}
\label{sec:asyncgetsmomenum}
Similar to Spike Compensation, \citeauthor{Mitliagkas2016AsynchronyBM}~\citeyearpar{Mitliagkas2016AsynchronyBM} also adjust the momentum for asynchronous training. They propose modifying the momentum coefficient to keep the ``average age" of the velocity the same as without a delay. SC instead applies the missing contribution of the delayed gradients immediately which is generally not equivalent to modifying or tuning the momentum coefficient. We experimented with the method suggested by \citeauthor{Mitliagkas2016AsynchronyBM}~\citeyearpar{Mitliagkas2016AsynchronyBM} but it barely changes the value of the momentum coefficient in our setting and did not aid training. One major difference is that the delays in our setting are fixed but \citeauthor{Mitliagkas2016AsynchronyBM}~\citeyearpar{Mitliagkas2016AsynchronyBM} assume an exponentially distributed gradient delay. For the convex quadratic, we also found that tuning the momentum coefficient is not sufficient to restore the effects of momentum in the constant delay setting.


\section{Simulating Delayed Gradients}
\label{pytorch_del_grad_sim}
Weight inconsistency and delayed gradients are potential issues in Pipelined Backpropagation. To better understand the issues we simulated weight inconsistency and delayed gradients in a PyTorch~\cite{paszke2017automatic} environment using a modified optimizer. The modified optimizer has a buffer of old parameter values. 
To apply a delay $D$, the model is loaded with parameters from $D$ time steps ago, a forward and backward pass is performed. 
The resulting gradients are then used to update a master copy of the weights. 
Weight inconsistency is simulated by loaded the model with parameters from $D$ time steps ago, doing the forward pass then loading the model with the master weights before doing the backwards pass. 
While this was not an exact model of PB, this setup allows for the simulation of PB's issues and fast iterate of potential methods to overcome the issues. This technique can also be used to simulate PB by having different delays for different layers based on the depth of the layer.
This simulation method does not allow simultaneously launching multiple kernels and is therefore not efficient for small batch sizes.
Our simulations are done using a constant delay across layers. This upper bounds the effect of weight inconsistency and delayed gradients. This setup can also be used to simulated ASGD training by making $D$ a random variable which models the distribution of GPU communications with the master node in ASGD.

\begin{table*}
    \caption{
        CIFAR-10 validation accuracy (mean$\pm$std.dev of five runs) comparing LWP$^\mathrm{v}_\mathrm{D}$ and LWP$^\mathrm{w}_\mathrm{D}$ on ResNet (RN) and VGG training.
    }
    \label{tab:c10accVvsW}
    \centering
    \vskip 0.15in
    \small
    \sc
    \begin{tabular}{l|cc|cc}
        \toprule
        Network & SGDM                    & PB                      & PB+LWP$^\mathrm{v}_\mathrm{D}$+SC\textsubscript{D} & PB+LWP$^\mathrm{w}_\mathrm{D}$+SC\textsubscript{D}  \\ \midrule
        VGG11   & \textbf{91.16}$\pm$0.19 & 90.83$\pm$0.20          & \textbf{91.12}$\pm$0.18                            & 90.93$\pm$0.15                                      \\
        VGG13   & \textbf{92.57}$\pm$0.15 & \textbf{92.59}$\pm$0.15 & \textbf{92.56}$\pm$0.14                            & \textbf{92.55}$\pm$0.08                             \\
        VGG16   & \textbf{92.24}$\pm$0.19 & 92.06$\pm$0.21          & \textbf{92.38}$\pm$0.27                            & 92.09$\pm$0.10                                      \\ \midrule
        RN20    & \textbf{90.63}$\pm$0.31 & 90.44$\pm$0.24          & \textbf{90.92}$\pm$0.25                            & \textbf{90.85}$\pm$0.41                             \\
        RN32    & \textbf{91.68}$\pm$0.23 & 91.46$\pm$0.09          & \textbf{92.04}$\pm$0.13                            & \textbf{91.99}$\pm$0.16                             \\
        RN44    & \textbf{92.19}$\pm$0.14 & 91.71$\pm$0.25          & \textbf{92.16}$\pm$0.26                            & \textbf{92.20}$\pm$0.36                             \\
        RN56    & \textbf{92.39}$\pm$0.20 & 91.89$\pm$0.40          & \textbf{92.48}$\pm$0.11                            & 92.32$\pm$0.06                                      \\
        RN110   & \textbf{92.77}$\pm$0.22 & 91.81$\pm$0.15          & \textbf{92.41}$\pm$0.16                            & 91.85$\pm$0.16                                      \\
        \bottomrule
    \end{tabular}
    \vskip -0.1in
\end{table*}

\begin{table*}[ht]
    \caption{
        CIFAR-10 final validation accuracy (mean$\pm$std.dev of 5 runs) for ResNet (RN) with group normalization.
    }
    \label{tab:c10accMeanRoundrtrPIPEMARE}
    \centering
    \vskip 0.15in
    \small
    \sc
    \resizebox{\textwidth}{!}{
    \begin{tabular}{l|ccccccc}
        \toprule
        Network & SGDM                    &  PB            & PipeDream      & PipeMare       & PipeMare       & PipeMare       & PB+LWP$^\mathrm{v}_\mathrm{D}$+SC\textsubscript{D} \\
                &                         &                &                & (K=12)         & (K=25)         & (K=50)         &                                                    \\ \midrule
        RN20    & 90.63$\pm$0.31          & 90.44$\pm$0.24 & 90.36$\pm$0.06 & 90.56$\pm$0.08 & 90.55$\pm$0.27 & 90.18$\pm$0.15 & \textbf{90.92}$\pm$0.25                            \\
        RN32    & 91.68$\pm$0.23          & 91.46$\pm$0.09 & 91.40$\pm$0.28 & 91.31$\pm$0.13 & 91.22$\pm$0.23 & 91.27$\pm$0.14 & \textbf{92.04}$\pm$0.13                            \\
        RN44    & \textbf{92.19}$\pm$0.14 & 91.71$\pm$0.25 & 91.72$\pm$0.14 & 91.79$\pm$0.29 & 91.52$\pm$0.15 & 91.53$\pm$0.35 & \textbf{92.16}$\pm$0.26                            \\
        RN56    & 92.39$\pm$0.20          & 91.89$\pm$0.40 & 91.82$\pm$0.19 & 91.98$\pm$0.08 & 91.79$\pm$0.18 & 91.61$\pm$0.22 & \textbf{92.48}$\pm$0.11                            \\ \bottomrule 
    \end{tabular}
    }
    \vskip -0.1in
\end{table*}


\section{Experiment Details}
\label{apdx:expdetails}

\subsection{VGG Experiments}
\label{apdx:vgg_exp_details}
\citeauthor{simonyan2014very}~\citeyearpar{simonyan2014very} do not provide a setup for training VGG on CIFAR-10. We adopt the VGG model, hyperparameters, and data preprocessing from \citeauthor{fu_2019}~\citeyearpar{fu_2019}.

\subsection{GProp validation}
\label{apdx:gprop_validation}

To validate our framework implementation, we compare batch parallel SGD, and fill \& drain SGD training.
We trained each setting, as well as the same network in PyTorch, 10 times to validate similar behavior. Figure \ref{fig:gprop_validation} shows the optimization of the different SGD training modes for the first 20 epochs. Numerical precision, network initialization, and data loading / augmentation randomness makes a numerical comparison for distinct runs impractical. Instead we show the mean and standard deviation of 10 runs. The different SGD modes in GProp are consistent and also match PyTorch's SGD convergence.

\subsection{ResNetv2}
\label{apdx:res_exp_details}
\citeauthor{he2016identity}~\citeyearpar{he2016identity} modified the original ResNet formulation given by \citeauthor{He_2016_CVPR}~\citeyearpar{He_2016_CVPR} by introducing the ResNet pre-activation block. We adopt the hyperparameters and data preprocessing from \citeauthor{Chiley2019OnlineNF}~\citeyearpar{Chiley2019OnlineNF}. 
Our experiments are done at batch size one where Batch Normalization is not effective.
We replace Batch Normalization with Group Normalization or Online Normalization.
For ImageNet ResNet-50 training, we used an initial group size of two as outlined in the Group Normalization paper.
\citeauthor{Wu_2018_ECCV}~\citeyearpar{Wu_2018_ECCV} do not tune Group Normalization for CIFAR-10 training. 
We use the same initial group size of two for our CIFAR-10 experiments.
For our Online Normalization experiments we use the default forward and backward decay factors.

\subsection{LWP\texorpdfstring{$^\mathrm{v}_\mathrm{D}$}{} vs LWP\texorpdfstring{$^\mathrm{w}_\mathrm{D}$}{}}
\label{apdx:VvsW_WP}

Table~\ref{tab:c10accVvsW} shows the results of using the two variants of LWP. When combined with SC, LWP$^\mathrm{v}_\mathrm{D}$ outperforms LWP$^\mathrm{w}_\mathrm{D}$. When the weight form is used the most recent gradient has a large effect on the velocity estimate used for the weight prediction.
For small batch sizes this approximation might be noisy decreasing the effectiveness of LWP.
A similar effect can be observed for LWP in general (Appendix~\ref{apdx:momentum_scaling_effect}) when very small momentum values are used which also leads to noisy predictions.


\section{Comparison with PipeMare}
\label{sec:pipemare}
PipeMare \cite{yang2019pipemare} is a recent work on stabilizing Pipelined Backpropagation.
They propose two methods: Discrepancy Correction which approximates the weight stashing from PipeDream with smaller memory overhead, and learning rate rescheduling (LR\_R) which is a specific type of learning rate warm-up which accounts for the delay of each stage.
Table~\ref{tab:c10accMeanRoundrtrPIPEMARE} compares PipeDream and PipeMare to our methods. To avoid having to tune the hyperparameter for Discrepancy Correction we instead combine learning rate rescheduling with weight stashing to approximate PipeMare. We show results for three choices of the K hyperparameter of LR\_R (shown in epochs). As can be seen LR\_R can help slightly in our setting but unlike our methods, it is unable to recover the baseline accuracy.

Unlike prior work on Pipelined Backpropagation we focus on a batch size of one. Although our methods do not strictly require a batch size of one, the small batch size has several advantages for hardware both hardware (Section~\ref{sec:fgpponHW}) and convergence (Section~\ref{sec:small_batch_sizes_and_online_normalization}). Smaller batch sizes decrease the effect of gradient delay and weight inconsistency (Figure~\ref{fig:c10r20PBbsizeacc}) making it easier to stabilize training. Smaller batch sizes also change the relative significance of gradient delay and weight inconsistency (Appendix~\ref{sec:InconsistencyVsDelay}). With large batch sizes, weight inconsistency is the main issue but in our setting gradient delay becomes more significant. Methods designed for the larger batch size setting, such as PipeDream, PipeMare, and Gradient Shrinking, naturally tend to focus on addressing weight inconsistency rather than the delay which is likely why those methods do generally not seem to perform very well in our setting. Likewise, our methods may not be optimal for the long delays and large weight inconsistencies that can arise with fine-grained PB using bigger batches. For the fine grained small batch setting surveyed in this work, SpecTrain is the only existing method we found to work well in our setting although it was not always able to fully recover SGDM accuracy (Table~\ref{tab:c10accSTWP}).

\end{document}